\documentclass[sn-mathphys, iicol]{sn-jnl}

\usepackage{times}
\usepackage{epsfig}
\usepackage{graphicx}
\usepackage{amsmath}
\usepackage{amssymb}
\usepackage{arydshln}

\usepackage{comment}
\usepackage{subfigure}

\usepackage{booktabs}

\newcommand{\tabincell}[2]{\begin{tabular}{@{}#1@{}}#2\end{tabular}}

\usepackage{color}

\newcommand{\jc}[1]{\textcolor{black}{#1}}
\newcommand{\ad}[1]{\textcolor{black}{#1}}
\newcommand{\cvmadd}[1]{\textcolor{black}{#1}}
\newcommand{\tvcjadd}[1]{\textcolor{black}{#1}}
\newcommand{\tvcj}[1]{\textcolor{black}{#1}}

\jyear{2022}%

\theoremstyle{thmstyleone}%
\theoremstyle{thmstyletwo}%

\theoremstyle{thmstylethree}%

\begin{document}

\title[Jincen Jiang et al.]{Unsupervised Contrastive Learning with Simple Transformation for 3D Point Cloud Data}


\author[1]{\fnm{Jincen} \sur{Jiang}}\email{jinec@nwsuaf.edu.cn}

\author[2]{\fnm{Xuequan} \sur{Lu}}\email{xuequan.lu@deakin.edu.au}

\author[3]{\fnm{Wanli} \sur{Ouyang}}\email{wanli.ouyang@sydney.edu.au}

\author*[1]{\fnm{Meili} \sur{Wang}}\email{wml@nwsuaf.edu.cn}

\affil[1]{\orgdiv{College of Information Engineering}, \orgname{Northwest A\&F University}, \orgaddress{\country{China}}}

\affil[2]{\orgdiv{School of Information Technology}, \orgname{Deakin University}, \orgaddress{\country{Australia}}}

\affil[3]{\orgdiv{School of Electrical and Information Engineering}, \orgname{The University of Sydney}, \orgaddress{\country{Australia}}}




\abstract{Though a number of point cloud learning methods have been proposed to handle unordered points, most of them are supervised and require labels for training. By contrast,  unsupervised  learning of point cloud data has received much less attention to date. In this paper, we propose a simple yet effective approach for unsupervised point cloud learning. In particular, we identify a very useful transformation which generates a good contrastive version of an original point cloud. They make up a pair. After going through a shared encoder and a shared head network, the consistency between the output representations are maximized with introducing two variants of contrastive losses to respectively facilitate downstream classification and segmentation.  
To demonstrate the efficacy of our method, we conduct  experiments on three downstream tasks which are 3D object classification (on ModelNet40 and ModelNet10), shape part segmentation (on ShapeNet Part dataset) as well as scene segmentation (on S3DIS). Comprehensive results show that our unsupervised contrastive representation learning enables impressive outcomes in object classification and semantic segmentation. It generally outperforms current unsupervised methods, and even achieves comparable performance to supervised methods. \textit{Our source codes will be made publicly available.}}




\maketitle

\section{Introduction}\label{sec:introduction}

Point cloud, as an effective representation for 3D geometric data, has attracted noticeable attention recently. It has been used for learning based segmentation, classification, object detection, etc. Promising results have been achieved among those application fields. In this work, we focus on the use of 3D point clouds for classification and segmentation tasks. They respectively target to automatically recognize 3D objects and predict segment labels, which are crucial in multimedia computing, robotics, etc.

Most of existing methods for 3D point cloud analysis \cite{wu20153d,riegler2017octnet,wang2017cnn, su2015multi,li2020end,lyu2020learning, qi2017pointnet,qi2017pointnet++,li2018pointcnn} use annotated data for training. Nevertheless, annotation is time-consuming and costly, especially for a considerable amount of data. In the real world, it is particularly challenging to have annotated data for training all the time. 
Unsupervised learning is a good alternative \cite{achlioptas2018learning,yang2018foldingnet,han2019multi,zhao20193d
}. For example, Latent-GAN  \cite{achlioptas2018learning} used a deep architecture of Autoencoder (AE), and trained a minimal GAN in the AE's latent space for learning representations of point clouds. FoldingNet  \cite{yang2018foldingnet} proposed a new AE to get the codeword which can represent the high dimensional embedding point cloud, and the fully-connected decoder was replaced with the folding-based decoder. MAP-VAE \cite{han2019multi} conducted half-to-half predictions (splitting point cloud into a front half and a back half with several angles), and then combined them with global self-supervision to capture the geometry and structure of the point cloud. 3D-PointCapsNet \cite{zhao20193d} used the encoder-decoder structure, 
and concatenated the features from the encoder to form the point capsules. These methods usually employ the AE as the backbone, and often suffer from the curse of less quality representations. 
As a result, they may still induce less desired performance on downstream tasks (e.g. classification and segmentation).

Motivated by the above analysis, we propose an unsupervised representation learning  method which can facilitate the downstream 3D object classification and semantic segmentation. Our core idea is to maximize the agreement or consistency between the representations of the original point cloud and its transformed version (i.e. contrastive version). 

We simply utilize one transformation to generate the transformed version point cloud and pair it with the original ones. And then feed them into a shared base encoder network (e.g. former part of PointNet \cite{qi2017pointnet} with global feature), followed by a subsequent projection head network (e.g. latter part of PointNet: several mlp layers). The agreement maximization is imposed on the outputs of the projection head network, to facilitate the training efficiency and better preserve the rich representations output from the encoder. Since there are no labels involved in training, it is unsupervised representation learning for 3D point cloud data.

To validate our unsupervised method, we conduct experiments for the object classification task on ModelNet40 and ModelNet10, the shape part segmentation task on ShapeNet Part dataset, and the scene segmentation task on the S3DIS dataset. 
Extensive results show that our unsupervised contrastive representative learning enables impressive outcomes in terms of the three tasks. Our method generally outperforms state-of-the-art unsupervised techniques, and is even comparable to certain supervised counterparts.

The contributions of this paper are:
\begin{itemize}
    \item an unsupervised representation learning approach which is simple yet effective on 3D point cloud data,
    \item \jc{a simple transformation in generating a good contrastive version of an original point cloud, which is better than other complex transformations,}
    \item two variants of point cloud based 
    contrastive losses for downstream classification and segmentation, respectively, 
    \item \tvcj{experiments and analysis on three tasks (classification, shape part segmentation and scene segmentation), achieving promising performance with our simple contrastive learning. }
\end{itemize}

\section{Related Work}
\label{sec:relatedwork}

\tvcj{Unlike 2D images, which consist of regular and uniform pixels, point cloud data are often irregular, sparse and contaminated with noise/outliers during the obtaining procedure of scanning and processing \cite{Zhang2020,LUdening2020,lu2020,lu2017}. 
3D point cloud learning techniques can be generally classified into three categories: (1) voxel based \cite{wu20153d,riegler2017octnet,wang2017cnn}, (2) view based \cite{su2015multi,su2018splatnet,zhou2019multi,li2020end,lyu2020learning} and (3) point based \cite{qi2017pointnet,qi2017pointnet++,li2018pointcnn,wu2019pointconv,xu2018spidercnn,liu2019relation,komarichev2019cnn,wang2019dynamic,lin2020convolution,jiang2021guided,du2021self}. Voxel based methods often involve resolution and memory issues, and view based approaches are often criticized for the tedious pre-processing, i.e. projecting each 3D object onto 2D image planes. Point based techniques are capable of learning features from point cloud data straightforwardly. In fact, most of these methods are supervised. }

\textbf{Voxel based techniques.} 3D volumetric CNNs (Convolutional Neural Network) imitates classical 2D CNNs by performing voxelization on the input point cloud. 3D ShapeNets was designed for learning volumetric shapes \cite{wu20153d}. Riegler et al. proposed OctNet for deep learning with sparse 3D data \cite{riegler2017octnet}. Wang et al. presented an Octree-based CNN for 3D shape analysis, which was called O-CNN \cite{wang2017cnn}. These methods are proposed to improve 3D volumetric CNNs and reach high volume resolutions.

\textbf{View based methods.} 
View based methods are to project 3D point cloud data onto the regular image planes. For example, MVCNNs used multiple images rendered from the 3D shapes to fit classical 2D CNNs \cite{su2015multi}. Su et al. proposed to utilize a sparse set of samples in a high-dimensional lattice as the representation of a collection of points \cite{su2018splatnet}. 
\ad{Zhou et al. proposed the multi-view saliency guided deep neural network (MVSG-DNN) which contains three modules to capture and extract the features of individual views to compile 3D object descriptors for 3D object retrieval and classification \cite{zhou2019multi}. Xu et al. used a LSTM-based network to recurrently aggregate the 3D objects shape embedding from an image sequence and estimate images of unseen viewpoints, aiming at the fusion of multiple views' features \cite{xu2019learning}. Huang et al. devised a view mixture model (VMM) to decompose the multiple views into a few latent views for the descriptor construction  \cite{huang2020learning}. }
Li et al. presented an end-to-end framework to learn local multi-view descriptors for 3D point clouds \cite{li2020end}. Lyu et al. projected 3D point clouds into 2D image space by learning the topology-preserving graph-to-grid mapping \cite{lyu2020learning}.

\textbf{Point based methods.} PointNet is a seminal work on point based learning \cite{qi2017pointnet}. In PointNet, max-pooling operation is used to learn permutation-invariant features. The original authors introduced PointNet++, a hierarchical neural network that applied PointNet recursively on a nested partitioning of the input point set \cite{qi2017pointnet++}. It achieved better learning outcomes than PointNet. Later, pointCNN was introduced to learn an X-transformation from the input points, to promote the weighting of the input features and the permutation of the points into a latent order \cite{li2018pointcnn}. PointConv, a density re-weighted convolution, was proposed to fully approximate the 3D continuous convolution on any set of 3D points \cite{wu2019pointconv}. Xu et al. proposed SpiderCNN to extract geometric features from point clouds \cite{xu2018spidercnn}. Liu et al. designed a Relation-Shape Convolutional Neural Network to learn the geometric topology constraints among points \cite{liu2019relation}. Simonovsky et al. generalized the convolution operator from regular grids to arbitrary graphs and applied it to point cloud classification \cite{simonovsky2017dynamic}. Parametric Continuous Convolution was introduced to exploit parameterized kernel functions that spanned the full continuous vector space \cite{wang2018deep}. Li et al. came up with a self-organizing network which applied hierarchical feature aggregation using self-organizing map \cite{li2018so}. It included a point cloud auto-encoder as pre-training
to improve network performance. Komarichev et al. presented an annular convolution operator to better capture the local neighborhood geometry of each point by specifying the (regular and dilated) ring-shaped structures and directions in the computation \cite{komarichev2019cnn}. Zhao et al. put forwarded PointWeb to enhance local neighborhood features for point cloud processing \cite{zhao2019pointweb}. Xie et al. developed a new representation by adopting the concept of shape context as the building block and designed a model (ShapeContextNet) for point cloud recognition \cite{xie2018attentional}. Wang et al. designed a new neural network module dubbed EdgeConv which acts on graphs dynamically computed in each layer \cite{wang2019dynamic}. More recently, Fujiwara et al. proposed to embed the distance field to neural networks \cite{fujiwara2020neural}. Lin et al. defined learnable kernels with a graph max-pooling mechanism for their 3D Graph Convolution Networks (3D-GCN) \cite{lin2020convolution}. Yan et al. presented the adaptive sampling and the local-nonlocal modules for robust point cloud processing \cite{yan2020pointasnl}. 
\ad{Qiu et al. proposed a network considering both low-level geometric information of 3D space points explicitly and high-level local geometric context of feature space  implicitly \cite{qiu2021geometric}. Chen et al. presented a hierarchical attentive pooling graph network (HAPGN) for segmentation which includes the gated graph attention network to get a better representation of local features and hierarchical graph pooling module to learn hierarchical features  \cite{chen2020hapgn}. Liu et al. devised a point context encoding module  (PointCE) and a semantic context encoding loss (SCE-loss) to capture the rich semantic context of a point cloud adaptively, achieving improved segmentation performance  \cite{liu2020semantic}. }

\begin{figure*}[htbp]
\centering
\begin{minipage}[b]{0.95\linewidth}
{\label{}
\includegraphics[width=1\linewidth]{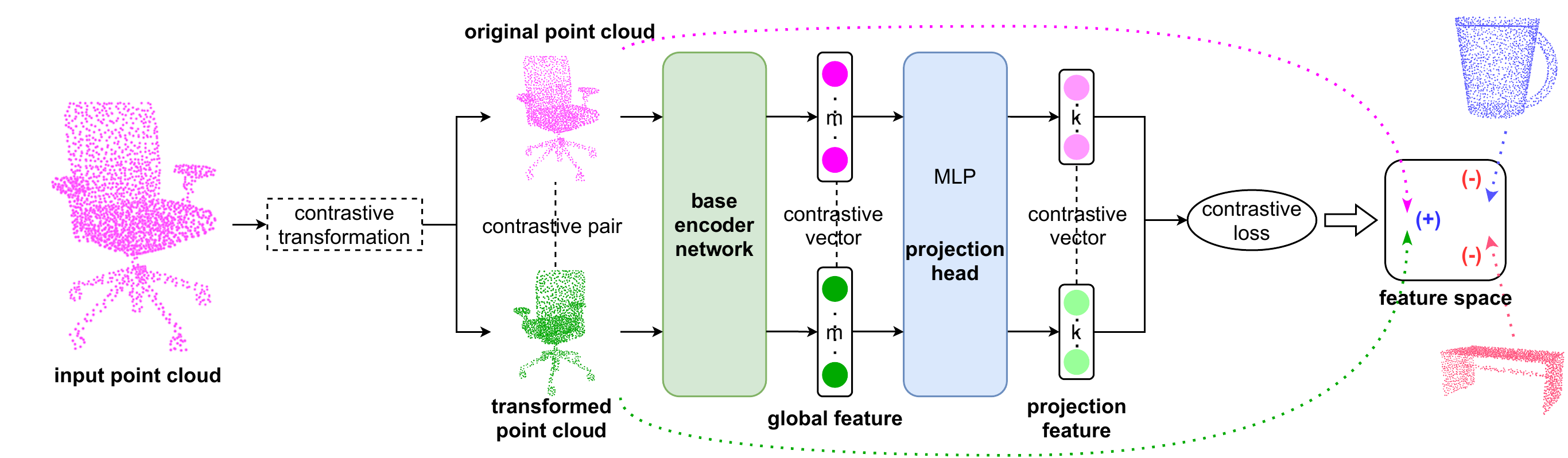}}
\end{minipage}
\caption{Overview of our unsupervised contrastive representation learning method. Given a point cloud, the transformation (rotation in this work) is used to the get transformed version of  the original point cloud, which defines a pair. 
Then, the pairs are input to the base encoder network (e.g. PointNet or DGCNN) to learn the global feature of each model. The projection head is further used to reduce the global feature dimension and for effective loss degradation. The contrastive loss encourages a pair of point clouds to be consistent in the feature space.
}
\label{fig:overview}
\end{figure*}

\textbf{Unsupervised representation learning.}
Yang et al. proposed an autoencoder (AE), referred to as FoldingNet, for unsupervised learning on point cloud data \cite{yang2018foldingnet}.   MAP-VAE was proposed to enable the learning of global and local geometry by jointly
leveraging global and local self-supervision \cite{han2019multi}. Rao et al. presented bidirectional reasoning between the local structures and the global 
shape for unsupervised representation learning of point clouds \cite{rao2020global}. It used a much larger RSCNN as backbone (4$\times$RSCNN) \cite{liu2019relation}. 
\ad{Zhang et al. presented an explainable machine learning method for point cloud classification by building local-to-global features through iterative one-hop information exchange, and feeding the feature vector to a random forest classifier for classification  \cite{zhang2020pointhop}. } 
Different from them, we create a contrastive pair for each point cloud, and our framework simply consists of an encoder network and a head network. The encoder outputs global representations (features) for downstream networks and the head outputs projection features (a smaller size) for calculating the loss. 

More recently, Xie et al. presented an unsupervised pre-training framework called PointContrast for high-level scene understanding tasks  \cite{xie2020pointcontrast}. Their findings demonstrated that the learned representation could generalize across domains.  \cite{xie2020pointcontrast} focused on 3D scenes (pretrained on a very large-scale generated dataset (about $1$ terabyte), and sophisticatedly considered matched points (i.e. common points) of two different views (at least 30\% overlap) as pairs. 
\jc{Unlike that, our point cloud level based approach simply uses a rotational transformation to generate a transformed version of an original point cloud. It can easily get a great pose discrepancy, without requiring point cloud overlap to satisfy the demand of obtaining a certain number of matched points.}
\jc{In essence, a pair of matched points are treated as a pair in  \cite{xie2020pointcontrast} to learn point-level features, while a pair of point clouds (a point cloud consisting of a series of points) are regarded as a pair in our work. Treating the point clouds as the pair in our method has the advantage of learning better global representations when compared with  \cite{xie2020pointcontrast}. It is also intuitive and straightforward to use point cloud level, while PointContrast \cite{xie2020pointcontrast} can hardly obtain point cloud representations directly and is suitable for point-wise tasks, e.g., scene segmentation. In comparison, our global feature of point cloud level can be easily used in both point cloud level and point-wise tasks (e.g., classification and segmentation).} 
Meanwhile, for unsupervised learning we derive two variants of contrastive  
losses based on point clouds,  which respectively facilitate two different types of downstream tasks (i.e., classification and segmentation). Finally, the backbone networks are different: they use a Sparse Residual U-Net \jc{which requires voxelization of point clouds}, while we use a simple encoder-head structure.

\tvcj{Difference from SimCLR \cite{chen2020simple}: It is derived on top of SimCLR \cite{chen2020simple}, but is largely different from SimCLR: (1) SimCLR is designed for 2D images, and our contrastive learning is for 3D point cloud data, which involves irregular point distribution, and poses more challenges than 2D images with regular grids. (2) SimCLR uses multiple transformations for an original image, and these two different versions of images act as a pair. By contrast, we only generate a transformed version of the original point cloud with a simple transformation (e.g. rotation), thus forming a contrastive pair of this point cloud (i.e. a pair in point cloud level). (3) SimCLR typically uses TPU resources with a large batch size for training, while we demonstrate an elegant approach for 3D point cloud representation learning, which is simple yet effective, without requiring TPU computing resources.
}

\section{Method}
\label{sec:method}

In this work, we take 3D object classification and semantic segmentation (shape and scene) as the downstream tasks of our unsupervised contrastive representation learning. To clearly elaborate our method, we take downstream object classification as an example when designing the unsupervised stage, and we will later explain how to extend it to shape and scene segmentation.

Given an unlabeled point cloud, we first use a transformation (i.e. rotation) to generate its transformed version, thus constructing a contrastive point cloud pair for this original point cloud. They are fed into a base encoder network, in order to learn a pair of global features or representations. The global features are then passed to a projection head network, to obtain another pair of representations. These two representations from the projection head network target to reach a maximum agreement with the aid of a contrastive loss function.  Figure \ref{fig:overview} shows the framework of our unsupervised contrastive representation learning.

\subsection{Unsupervised Contrastive Representation Learning}
\label{sec:unsupervisedcontrastive}
\textbf{Contrastive transformation.} 
Unlike 2D images, point cloud data often have an irregular distribution in 3D space, and have a complex degree of freedom. Given this, it is more difficult to identify the practically useful transformations for constructing a good contrastive pair of a point cloud. 
\tvcj{SimCLR \cite{chen2020simple} utilized different types of transformations for a single image (e.g. cropping, rotation), and generated two transformed versions. In contrast, we only use one transformation for simplicity, and pair the original point cloud with the transformed version, that is, a pair including the original point cloud and its transformed counterpart.} Common transformations in 3D space are \jc{rotation, cutout, crop, scaling, smoothing,  noise corruption, etc.} Since heavy noise corruption will destroy the object shapes, we exclude this for transformations applied here.  Jittering is analogous to light noise, and we use jittering for data augmentation, following the protocol of the state-of-the-art point based methods. In this work, we select rotation as the transformation, and use it to generate a transformed version for the original point cloud. We provide the discussion of the choice in ablation studies (Section \ref{sec:ablation}).

\textbf{Base encoder network.} 
Point based networks, such as PointNet \cite{qi2017pointnet}, DGCNN \cite{wang2019dynamic}, Pointfilter \cite{Zhang2020}, often involve a pooling layer to output the global feature for an input point cloud. The former part of a point based network before this layer (inclusive) can be naturally viewed as a based encoder in our framework. In other words, the input point cloud can be encoded into a latent representation vector (i.e. the global feature). In this sense, we can simply extract this former part of any such point based networks as a base encoder network in our unsupervised contrastive representation framework. In this work, we select some state-of-the-art point based networks including PointNet and DGCNN as the backbone, and extract their former parts as our base encoder accordingly. It is interesting to discover that the encoders involving T-Net (i.e. transformation net) will hinder the learning of unsupervised contrastive representations. We deduce that T-Net accounts for various rotated point cloud augmentations, which degrades the ability of capturing a large contrast between the input pair. As such, we remove the original T-Net (i.e. transformation net) in these encoders, if involved. We show the results of different encoders in Section \ref{sec:results}.

\textbf{Projection head network.} 
Point based networks usually have several fully connected layers to bridge the global feature with the final $k$-class vector. Similar to the encoder, we can also simply extract the latter part of a point based network as the projection head. Alternatively, it is also flexible to customize a projection head network by designing more or fewer fully connected layers.

Mathematically, the final $k$-class vector (or representation vector) can be formulated as
\begin{equation}\label{eq:kvector}
\begin{aligned}
\mathbf{z_i} &= H(E(\mathbf{P})), \\
\mathbf{z_j} &= H(E(\mathbf{P}')),
\end{aligned}
\end{equation}
where $\mathbf{P}$ is an original point cloud and $\mathbf{P}'$ is its transformed counterpart. $E$ and $H$ denote the encoder network and the projection head network, respectively.

\textbf{Contrastive loss function.} 
We first randomly select $n$ samples, and use the selected transformation (i.e. rotation) to generate another $n$ corresponding transformed counterparts, resulting in $n$ pairs ($2n$ samples) constituting the minibatch. Analogous to SimCLR \cite{chen2020simple}, we also do not explicitly define positive or negative pairs. Instead, we select a pair as the positive pair, and the remaining $(n-1)$ pairs (i.e. $2(n-1)$ samples) are simply regarded as negative pairs.

As for the unsupervised loss function, InfoNCE \cite{oord2018representation} is a widely-used loss function for unsupervised representation learning of 2D images. More recently, \cite{xie2020pointcontrast} also utilized a similar loss for contrastive scene representation learning. Inspired by them, we also introduce a variant as our unsupervised loss function, which is defined as

\begin{equation}\label{eq:contrastiveloss}
\begin{aligned}
L = - \frac{1}{|$S$|}\sum_{(i, j)\in S} \log{\frac{\exp(\mathbf{z_i}\cdot\mathbf{z_j}/\tau)}{\sum_{(\cdot, t)\in S, t \neq j}\exp(\mathbf{z_i}\cdot\mathbf{z_t}/\tau)}},
\end{aligned}
\end{equation}
where $S$ is the set of all positive pairs (point cloud level), and $\tau$ is a temperature parameter. $||$ denotes the cardinality of the set. The loss is computed using all the contrastive pairs, and is equivalent to applying the cross entropy with pseudo labels (e.g. $0\sim15$ for $16$ pairs). We found it works very well in our unsupervised contrastive representation learning.

\subsection{Downstream 3D Object Classification}
\label{sec:method-classificaion}
We take 3D object classification as the first downstream task in this work, to validate our unsupervised representation learning. The above designed scheme is immediately ready for the unsupervised representation learning to facilitate the downstream classification task. 
In particular, we will utilize two common schemes for validation here. One is to train a linear classification network by taking the learned representations of our unsupervised learning as input. Here, the learned representation is the global feature. We did not choose the $k$-class representation vector as it had less discriminative features than the global feature in our framework, and it induced  a poor performance (see Section \ref{sec:ablation}). The other validation scheme is to initialize the backbone with the unsupervised trained model and perform a supervised training. 
We will demonstrate the classification results for these two validation schemes in Section \ref{sec:results}.

\subsection{Downstream Semantic Segmentation}
\label{sec:semanticsegmentation}

\tvcj{To further demonstrate the effectiveness of our unsupervised representation learning, we also fit the above unsupervised learning scheme to the downstream semantic segmentation, including shape part segmentation and scene segmentation. 
Since it is a different task from 3D object classification, we need to design a new scheme to facilitate  unsupervised training. We still use the rotation to generate a transformed version of an original point cloud (e.g. a shape point cloud or a split block from the scene), and view them as a contrastive pair (i.e. point cloud level). As for segmentation, each point in the point cloud has a feature representation. For unsupervised   representation learning, we compute the mean of all point-wise cross entropy to evaluate the overall similarity within the mini-batch. 
We therefore define a loss function for semantic segmentation as:
} 
\begin{equation}
\label{eq:contrastiveloss_seg}
\begin{aligned}
L = & - \frac{1}{S}\sum_{(a, b)\in S} \frac{1}{P_{(a, b)}}\sum_{(i, j)\in P_{(a, b)}} \\
&\log{\frac{\exp(\mathbf{z_i}\cdot\mathbf{z_j}/\tau)}{\sum_{(\cdot, t)\in P_{(a, b)}, t \neq j}\exp(\mathbf{z_i}\cdot\mathbf{z_t}/\tau)}},
\end{aligned}
\end{equation}

where $S$ is the set of all positive pairs (i.e. point cloud $a$ and $b$), and $P_{(a, b)}$ is the set of all point pairs (i.e. the same point id) of the point cloud $a$ and $b$. Similarly, we apply the cross entropy with pseudo labels which match the point indices (e.g. $0\sim2047$ for $2048$ points).

\section{Experimental Results}
\label{sec:results}

\begin{table*}[htbp]
    \centering
    \caption{ \tvcj{Classification results of unsupervised methods and our method (Linear Classifier), on the datasets of ModelNet40 and ModelNet10. Both ShapeNet55 (upper part) and ModelNet40 (bottom part) pretrained datasets are provided.}
    }\label{table:classification_unsupervised}
    \begin{tabular}{l c c c c c}
        \hline
         \tabincell{c}{Methods} &  \tabincell{c}{Pretrained Dataset} &  \tabincell{c}{Input Data} & \tabincell{c}{Resolution\\e.g. \# Points} &  \tabincell{c}{ModelNet40\\ Accuracy} &  \tabincell{c}{ModelNet10\\Accuracy} \\ 
         \hline
         Latent-GAN \cite{achlioptas2018learning} & ShapeNet55 & xyz & 2k & 85.70 & 95.30\\
         FoldingNet \cite{yang2018foldingnet} & ShapeNet55 & xyz & 2k & 88.40 & 94.40\\
         MRTNet \cite{gadelha2018multiresolution} & ShapeNet55  & xyz & multi-resolution & 86.40 & - \\
         3D-PointCapsNet \cite{zhao20193d} & ShapeNet55 & xyz & 2k & 88.90 & - \\
        \hline
        \jc{Ours (DGCNN)} & ShapeNet55 & xyz & 2k & \jc{89.37} & -\\
        \hline
         VIPGAN \cite{han2019view} & ModelNet40 & views & 12 & 91.98 & 94.05\\
         Latent-GAN \cite{achlioptas2018learning} & ModelNet40 & xyz & 2k & 87.27 & 92.18\\
         FoldingNet \cite{yang2018foldingnet} & ModelNet40 & xyz & 2k & 84.36 & 91.85\\
         3D-PointCapsNet \cite{zhao20193d} & ModelNet40 & xyz & 1k & 87.46 & - \\
         \ad{PointHop} \cite{zhang2020pointhop} & ModelNet40 & xyz & 1k & 89.10 & - \\
         MAP-VAE \cite{han2019multi} & ModelNet40 & xyz & 2k & 90.15 & 94.82\\
         GLR (RSCNN-Large) \cite{rao2020global} & ModelNet40 & xyz & 1k & 92.9 & - \\
         \hline
         Ours (PointNet \cite{qi2017pointnet}) & ModelNet40 & xyz & 1k & 88.65 & 90.64\\
         Ours (DGCNN \cite{wang2019dynamic}) & ModelNet40 & xyz & 1k & 90.32 & \jc{95.09}\\
         \hline
    \end{tabular}
\end{table*}

\subsection{Datasets}

\textbf{Object classification.}
We utilize ModelNet40 and ModelNet10 \cite{wu20153d} for 3D object classification. We follow the same data split protocols of PointNet-based methods \cite{qi2017pointnet, qi2017pointnet++, wang2019dynamic} for these two datasets. For ModelNet40, the train set has $9,840$ models and the test set has $2,468$ models, and the datset consists of $40$ categories. For ModelNet10, $3,991$ models are for training and $908$ models for testing. It contains $10$ categories. For each model, we use $1,024$ points with only $(x,y,z)$ coordinates as the input, which is also consistent with previous works.

\jc{Note that some methods \cite{yang2018foldingnet,gadelha2018multiresolution,achlioptas2018learning,zhao20193d} are pre-trained under the ShapeNet55 dataset \cite{chang2015shapenet}. We also conduct a version of ShapeNet55 training for the classification task. We used the same dataset as \cite{zhao20193d}, which has $57,448$ models with $55$ categories, and all models will be used for unsupervised training. Following the same setting of previous work, we use $2,048$ points as input. }

\jc{We provide comparison experiments with PointContrast \cite{xie2020pointcontrast} for the classification task, and they use the ShapeNetCore \cite{chang2015shapenet} for finetuning. The dataset contains $51,127$ pre-aligned shapes from $55$ categories, which has $35,708$ models for training, $5,158$ models for validation and $10,261$ models for testing. We use $1,024$ points as input which is the same as PointContrast \cite{xie2020pointcontrast}.}

\textbf{Shape part segmentation.} 
We use the ShapeNet Part dataset \cite{yi2016scalable} for shape part segmentation, which consists of $16,881$ shapes from $16$ categories. Each object involves 2 to 6 parts, with a total number of $50$ distinct part labels.  We follow the official dataset split and the same point cloud sampling protocol as \cite{chang2015shapenet}. Only the point coordinates are used as input. Following \cite{qi2017pointnet++,wang2019dynamic}, we use mean Intersection-over-Union (mIoU) as the evaluation metric.

\textbf{Scene segmentation.}
We also evaluate our model for scene segmentation on Stanford Large-Scale 3D Indoor Spaces Dataset (S3DIS) \cite{armeni2017joint}. This dataset contains 3D scans of $271$ rooms and $6$ indoor areas, covering over $6,000 m^2$.  We follow the same setting as \cite{qi2017pointnet++, wang2019dynamic}. Each room is split with $1m \times 1m$ area into little blocks, and we sampled $4,096$ points of each block. Each point is represented as a 9D vector, which means the point coordinates, RGB color and normalized location for the room. Each point is annotated with one of the $13$ semantic categories. We also follow the same protocol of adopting the six-fold cross validation for the six area.

\textit{Please refer to the \jc{Appendices} for additional information and visual results. }

\subsection{Experimental Setting}
We use Adam optimizer for our unsupervised representation training. We implemented our work with TensorFlow, and use a single TITAN V GPU for training (DGCNN using multiple GPUs).

For downstream 3D object classification on ModelNet40, ModelNet10, \jc{ShapeNet55 and ShapeNetCore}, we use a batch size of $32$ (i.e. $16$ contrastive pairs) for training and testing. Temperature hyper-parameter $\tau$ is set as $1.0$. We use the same dropouts with the original methods accordingly, i.e. $0.7$ for PointNet as backbone, $0.5$ for DGCNN as backbone. The initial decay rate of batch normalization is $0.5$, and will be increased no lager than $0.99$. The training starts with a $0.001$ learning rate, and is decreased to $0.00001$ with an exponential decay.

We employ DGCNN as the backbone for semantic segmentation. As for shape part segmentation on ShapeNet Part dataset, we utilize a batch size of $16$ (i.e. $8$ constrasive pairs) for training. We use a batch size of $12$ (i.e. $6$ constrasive pairs) for scene segmentation on S3DIS. For the two tasks, we simply use a batch size of $1$ during testing, and the other settings follow DGCNN.

\subsection{3D Object Classification}

We conduct two kinds of experiments to evaluate the learned representations of our unsupervised contrastive learning. We first train a simple linear classification network with the unsupervised representations as input. Secondly, we take our unsupervised representation learning as pretraining, and initialize the weights of the backbone before supervised training. \cvmadd{Tables \ref{table:classification_unsupervised} and \ref{table:classification_supervised} show 3D object classification results for our method and a wide range of state-of-the-art techniques.}

\textbf{Linear classification evaluation.}
In this part, we use the former part of PointNet \cite{qi2017pointnet} and DGCNN \cite{wang2019dynamic} as the base encoder, and use the latter mlp layers as the projection head. 
The learned features are used as the input for training the linear classification network. We use the test accuracy as the evaluation of our unsupervised contrastive learning. Comparisons are reported in Table \ref{table:classification_unsupervised}. 
Regarding linear classification evaluation, our method with DGCNN as backbone always performs better than our method using PointNet as backbone, for example, $95.09\%$ versus $90.64\%$ for ModelNet10, $90.32\%$ versus $88.65\%$ for ModelNet40. This is due to a more complex point based structure of DGCNN. \cvmadd{Our method with DGCNN as backbone also outperforms most unsupervised techniques, like two recent methods PointHop ($1.22\%$ gain) and MAP-VAE ($0.17\%$ gain), and is comparable to some supervised methods in Table \ref{table:classification_supervised}, for example, $90.32\%$ versus $90.6\%$ (O-CNN) and $90.9\%$ (SO-Net with xyz) on ModelNet40, $95.09\%$ versus $93.9\%$ (supervised 3D-GCN) on ModelNet10.}  The GLR \cite{rao2020global} mined rich semantic and structural information, and used a larger RSCNN as backbone  (4$\times$RSCNN) \cite{liu2019relation}, resulting in a better accuracy than our method.

\jc{Notice that some methods used a larger ShapeNet55 dataset for training \cite{yang2018foldingnet,gadelha2018multiresolution,achlioptas2018learning,zhao20193d}.  Although the previous work \cite{han2019multi} re-implemented them by training on ModelNet40, they use $2048$ points rather than our $1024$. To provide additional insights, we re-implement and train a state-of-the-art method (3D-PointCapsNet \cite{zhao20193d}) on  ModelNet40 with $1024$ points, and train a linear classifier for evaluation. We choose this method since its code is publicly available and it is recent work. From Table \ref{table:classification_unsupervised}, it is obvious that our method (DGCNN as backbone) still  outperforms 3D-PointCapsNet by a $2.86\%$ margin.}

\jc{To show our learned representations have the transfer capability, we also train our method (DGCNN as backbone) on ShapeNet55 dataset for unsupervised contrastive learning and then feed the ModelNet40 dataset to the trained model to get point cloud features. 
We use these features to train a linear  classifier on ModelNet40 for evaluation. From Table \ref{table:classification_unsupervised} we can see that our method achieves the best result compared with other state-of-art methods on the same settings (Latent-GAN  \cite{achlioptas2018learning}, FoldingNet \cite{yang2018foldingnet}, and 3D-PointCapsNet \cite{zhao20193d}), exceeding them by $3.67\%$, $0.97\%$, and $0.47\%$, respectively. }

\tvcjadd{We also conduct an experiment for training with limited data to verify the capability of our pretrained model with linear classifier. The results can be seen in Table \ref{table:limited_data}. Our pretrained model achieves $86.6\%$ with $30\%$ data, which is $2.4\%$ higher compared with FoldingNet's \cite{yang2018foldingnet} $84.2\%$. With less data, e.g. $10\%$ of the data, our result is $0.6\%$ lower than FoldingNet. We suspect that with far fewer data, our contrastive learning method could less effectively learn the features of each sample with such simple transformation, and result in less improvement.  
}

\begin{table}[htb]
    \centering
    \caption{ Comparison results of classification accuracy with limited training data (different ratios).}
    \label{table:limited_data}
    \begin{tabular}{l c c c}
        \hline
         \tabincell{c}{Methods} &  10\% & 20\% & 30\%\\ 
        \hline
        FoldingNet \cite{yang2018foldingnet}) & 81.2 & 83.6 & 84.2\\
        Ours (DGCNN as backbone) & 80.6 & 85.4 & 86.6\\
        \hline
    \end{tabular}
\end{table}

\begin{table*}[htbp]
    \centering
    \caption{ \tvcj{Classification results of other supervised methods and our method (Pretraining), on the datasets of ModelNet40 and ModelNet10. We distinguish the results of other methods from our method by the line. 
    }
    }\label{table:classification_supervised}
    \begin{tabular}{l c c c c c}
        \hline
         \tabincell{c}{Methods} &  \tabincell{c}{Input Data} & \tabincell{c}{Resolution\\e.g. \# Points} &  \tabincell{c}{ModelNet40\\ Accuracy} &  \tabincell{c}{ModelNet10\\Accuracy} \\ 
        \hline
        Kd-Net (depth=10) \cite{klokov2017escape} & tree & $2^{10} \times3$ & 90.6 & 93.3\\
        PointNet++ \cite{qi2017pointnet++} & xyz & 1k & 90.7 & -\\
        KCNet \cite{shen2018mining} & xyz & 1k & 91.0 & 94.4 \\
        MRTNet \cite{gadelha2018multiresolution}  & xyz & 1k & 91.2 & - \\
        SO-Net \cite{li2018so} & xyz & 2k & 90.9 & 94.1\\
        KPConv \cite{thomas2019kpconv} & xyz & 6.8k & 92.9 & -\\
        PointNet++ \cite{qi2017pointnet++} & xyz, normal & 5k & 91.9 & -\\
        SO-Net \cite{li2018so} & xyz, normal & 5k & 93.4 & -\\
        O-CNN \cite{wang2017cnn} & xyz, normal & - & 90.6 & -\\
        PointCNN \cite{li2018pointcnn} & xyz & 1k & 92.2 & -\\
        PCNN \cite{atzmon2018point} & xyz & 1k & 92.3 & 94.9\\
        Point2Sequence \cite{liu2019point2sequence} & xyz & 1k & 92.6 & 95.3\\
        RS-CNN (voting) \cite{liu2019relation}  & xyz & 1k & 93.6 & -\\
        Neural Implicit \cite{fujiwara2020neural} & weights & $1024\times256$ & 92.2 & 95.7\\
        PointASNL \cite{yan2020pointasnl} & xyz & 1k & 92.9 & 95.7\\
        3D-GCN \cite{lin2020convolution} & xyz & 1k & 92.1 & 93.9\\
        \ad{HAPGN} \cite{chen2020hapgn} & xyz & 1k & 91.7 & -\\
        \ad{MVSG-DNN} \cite{zhou2019multi} & views & 12 & 92.3 & 94.0\\
        \hline
        Scratch (PointNet \cite{qi2017pointnet}) & xyz & 1k & 89.2 & - \\
        Ours (PointNet \cite{qi2017pointnet}) & xyz & 1k & 90.44 (+1.24) & 94.38\\
        \hdashline
        Scratch (DGCNN \cite{wang2019dynamic}) & xyz & 1k & 92.2 & - \\
        Ours (DGCNN \cite{wang2019dynamic})  & xyz & 1k & 93.03 (+0.83) & 95.93\\
        \hline
    \end{tabular}
\end{table*}

\textbf{Pretraining evaluation.} In addition to the above evaluation using a linear classifier, we further utilize the pre-training evaluation to demonstrate the efficacy of our unsupervised contrastive representation learning. Specifically, we also select PointNet and DGCNN as the backbone, in which the part before and including the global feature is regarded as the base encoder, and the remaining classification branch (i.e. several mlp layers) as the projection head. After our unsupervised representation training, we initialize the corresponding network with the unsupervised trained model, and then perform the supervised training. 
Table \ref{table:classification_supervised} shows the comparison results of our method and the state-of-the-art 3D object classification techniques with supervised training. 

\cvmadd{The pretraining evaluation based on our unsupervised representation learning  sees an improvement over the original backbone network, increased from $89.2\%$ to $90.44\%$ ($1.24\%$ increase) with PointNet and from $92.2\%$ to $93.03\%$ ($0.83\%$ increase) with DGCNN on ModelNet40.} Regarding ModelNet10, the accuracy of PointNet as our backbone for pretraining evaluation is $94.38\%$, which is on par with the supervised 3D-GCN ($93.9\%$). It is interesting to see that our method (DGCNN as backbone) is the best one on ModelNet10 in the pretraining evaluation, while the second best is achieved by two very recent supervised methods (Neural Implicit \cite{fujiwara2020neural} and PointASNL \cite{yan2020pointasnl}). \cite{fujiwara2020neural} even used a large weight matrix as the input for classification training. 
For ModelNet40, our method (DGCNN as backbone) achieves $93.03\%$, outperforming almost all techniques including both supervised and unsupervised ones. For example, our method in this case  outperforms the very recent supervised methods including 3D-GCN \cite{lin2020convolution}, Neural Implicit  \cite{fujiwara2020neural} and PointASNL \cite{yan2020pointasnl}. 

Compared to using PointNet as backbone, taking DGCNN as backbone achieves a better classification accuracy, for example, $95.93\%$ versus $94.38\%$, $93.03\%$ versus $90.44\%$. Similarly, we believe this is mainly because DGCNN exploits richer information than PointNet.

\jc{PointContrast \cite{xie2020pointcontrast} also presented an unsupervised contrastive learning approach, which is based on point level while ours is based on point cloud level. They validated their effectiveness on some datasets using the pretrain-finetuning strategy. In order to provide a potential comparison with it, we also used the ShapeNetCore dataset for the classification task with pretraining evaluation.
The comparison results are shown in Table  \ref{table:compare_pointcontrast}, and we can see that our method (DGCNN as backbone) outperforms them by $0.5\%$, though PointContrast is pretrained on a rather larger dataset (ScanNet). Note that our method is not suitable to be pretrained on ScanNet since this downstream task is for classification (requiring point cloud level features for classification) while ScanNet has point-wise labels. The good performance of our method is mainly due to the proper design of point cloud level based contrastive pairs and contrastive learning, so that we can directly obtain the global feature from  contrastive representation learning. We also re-implement DGCNN \cite{wang2019dynamic} on the ShapeNetCore dataset, which further demonstrates the effectiveness of our method by increasing from $84.0\%$ (original DGCNN) to $86.2\%$. In comparison with PointContrast which improved $0.6\%$ from the version of training from scratch, we achieve $2.2\%$ increase. }

\begin{table}[htb]
    \centering
    \caption{ Comparison results of PointContrast and our method on the dataset of ShapeNetCore with Pretraining evaluation. Note that * represents that the model is trained on ScanNet.}
    \label{table:compare_pointcontrast}
    \begin{tabular}{l c}
        \hline
         \tabincell{c}{Methods} &  \tabincell{c}{Accuracy}\\ 
        \hline
        Trained from scratch (PointContrast \cite{xie2020pointcontrast}) & 85.1\\
        PointContrast* \cite{xie2020pointcontrast} & 85.7\\
        Trained from scratch (original DGCNN \cite{wang2019dynamic}) & 84.0\\
        Ours (DGCNN as backbone) & 86.2\\
        \hline
    \end{tabular}
\end{table}

\subsection{Shape Part Segmentation}
\begin{figure*}[htb]\footnotesize
\centering
\begin{minipage}[b]{0.16\linewidth}
{\label{}\includegraphics[width=1\linewidth]{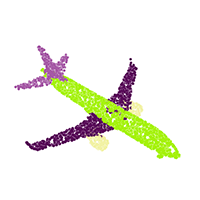}}
\centerline{airplane}
\end{minipage}
\begin{minipage}[b]{0.16\linewidth}
{\label{}\includegraphics[width=1\linewidth]{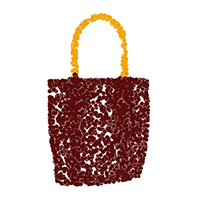}}
\centerline{bag}
\end{minipage}
\begin{minipage}[b]{0.16\linewidth}
{\label{}\includegraphics[width=1\linewidth]{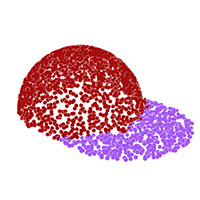}}
\centerline{cap}
\end{minipage}
\begin{minipage}[b]{0.16\linewidth}
{\label{}\includegraphics[width=1\linewidth]{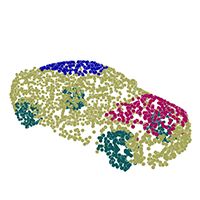}}
\centerline{car}
\end{minipage}
\begin{minipage}[b]{0.16\linewidth}
{\label{}\includegraphics[width=1\linewidth]{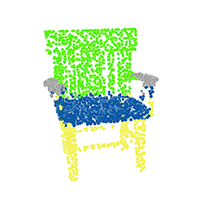}}
\centerline{chair}
\end{minipage}
\begin{minipage}[b]{0.16\linewidth}
{\label{}\includegraphics[width=1\linewidth]{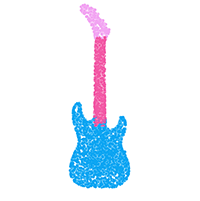}}
\centerline{guitar}
\end{minipage}\\
\begin{minipage}[b]{0.16\linewidth}
{\label{}\includegraphics[width=1\linewidth]{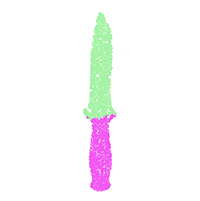}}
\centerline{knife}
\end{minipage}
\begin{minipage}[b]{0.16\linewidth}
{\label{}\includegraphics[width=1\linewidth]{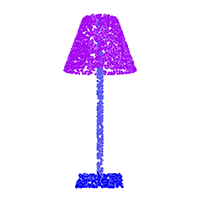}}
\centerline{lamp}
\end{minipage}
\begin{minipage}[b]{0.16\linewidth}
{\label{}\includegraphics[width=1\linewidth]{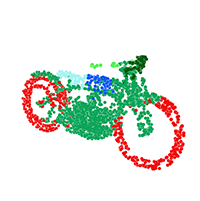}}
\centerline{motorbike}
\end{minipage}
\begin{minipage}[b]{0.16\linewidth}
{\label{}\includegraphics[width=1\linewidth]{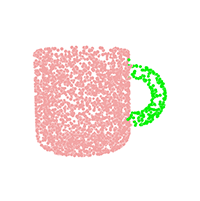}}
\centerline{mug}
\end{minipage}
\begin{minipage}[b]{0.16\linewidth}
{\label{}\includegraphics[width=1\linewidth]{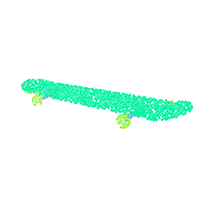}}
\centerline{skateboard}
\end{minipage}
\begin{minipage}[b]{0.16\linewidth}
{\label{}\includegraphics[width=1\linewidth]{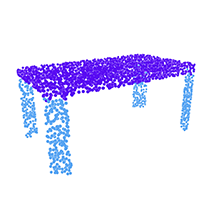}}
\centerline{table}
\end{minipage}
\caption{ Some examples of shape part segmentation using our method (Linear Classifier setting).  }
\label{fig:partseg}
\end{figure*}

In addition to the 3D object classification, we also verify our method on shape part segmentation. 
The segmentation results are listed in Table \ref{table:segmentation}. Here we take DGCNN as the backbone of our approach and simply employ the linear classifier evaluation setting. It can be seen from the table that our method in linear classification evaluation achieves  $79.2\%$ instance mIOU and $75.5\%$ class mIOU, which are remarkably better than state-of-the-art unsupervised techniques including MAP-VAE \cite{han2019multi} and Multi-task \cite{hassani2019unsupervised}. Specifically, our method outperforms MAP-VAE \cite{han2019multi} and Multi-task \cite{hassani2019unsupervised} by a margin of $7.55\%$ and $3.4\%$, respectively, in terms of class mIOU. 
Figure \ref{fig:partseg} illustrates some examples of our method (Linear Classifier setting) on the task of shape part segmentation.

\setlength{\tabcolsep}{3pt}
\begin{table*}[htb]\footnotesize
    \centering
    \caption{\tvcj{Shape part segmentation results of our method (Linear Classifier) and state-of-the-art techniques on ShapeNet Part dataset. We distinguish between supervised and unsupervised learning methods by the line. }
    }\label{table:segmentation}
    \begin{tabular}{l c c c c c c c c c c c c c c c c c c c}
        \hline
         Methods & Supervised & \tabincell{c}{class\\mIOU} & \tabincell{c}{instance\\mIOU} & air. & bag & cap & car & chair & ear. & guit. & kni. & lam. & lap. & mot. & mug & pist. & rock. & ska. & tab.\\
         \hline
        Kd-Net \cite{klokov2017escape} & yes & 77.4 & 82.3 & 80.1 & 74.6 & 74.3 & 70.3 & 88.6 & 73.5 & 90.2 & 87.2 & 81.0 & 84.9 & 87.4 & 86.7 & 78.1 & 51.8 & 69.9 & 80.3\\
MRTNet \cite{gadelha2018multiresolution} & yes & 79.3 & 83.0 & 81.0 & 76.7 & 87.0 & 73.8 & 89.1 & 67.6 & 90.6 & 85.4 & 80.6 & 95.1 & 64.4 & 91.8 & 79.7 & 87.0 & 69.1 & 80.6\\
PointNet \cite{qi2017pointnet} & yes & 80.4 & 83.7 & 83.4 & 78.7 & 82.5 & 74.9 & 89.6 & 73.0 & 91.5 & 85.9 & 80.8 & 95.3 & 65.2 & 93.0 & 81.2 & 57.9 & 72.8 & 80.6\\
KCNet \cite{shen2018mining} & yes & 82.2 & 84.7 & 82.8 & 81.5 & 86.4 & 77.6 & 90.3 & 76.8 & 91.0 & 87.2 & 84.5 & 95.5 & 69.2 & 94.4 & 81.6 & 60.1 & 75.2 & 81.3\\
RS-Net \cite{huang2018recurrent} & yes & 81.4 & 84.9 & 82.7 & 86.4 & 84.1 & 78.2 & 90.4 & 69.3 & 91.4 & 87.0 & 83.5 & 95.4 & 66.0 & 92.6 & 81.8 & 56.1 & 75.8 & 82.2\\
SO-Net \cite{li2018so} & yes & 81.0 & 84.9 & 82.8 & 77.8 & 88.0 & 77.3 & 90.6 & 73.5 & 90.7 & 83.9 & 82.8 & 94.8 & 69.1 & 94.2 & 80.9 & 53.1 & 72.9 & 83.0\\
PointNet++ \cite{qi2017pointnet++} & yes & 81.9 & 85.1 & 82.4 & 79.0 & 87.7 & 77.3 & 90.8 & 71.8 & 91.0 & 85.9 & 83.7 & 95.3 & 71.6 & 94.1 & 81.3 & 58.7 & 76.4 & 82.6\\
DGCNN \cite{wang2019dynamic} & yes & 82.3 & 85.2 & 84.0 & 83.4 & 86.7 & 77.8 & 90.6 & 74.7 & 91.2 & 87.5 & 82.8 & 95.7 & 66.3 & 94.9 & 81.1 & 63.5 & 74.5 & 82.6\\
KPConv \cite{thomas2019kpconv} & yes & 85.1 & 86.4 & 84.6 & 86.3 & 87.2 & 81.1 & 91.1 & 77.8 & 92.6 & 88.4 & 82.7 & 96.2 & 78.1 & 95.8 & 85.4 & 69.0 & 82.0 & 83.6\\
Neural Implicit \cite{fujiwara2020neural} & yes & - & 85.2 & 84.0 & 80.4 & 88.0 & 80.2 & 90.7 & 77.5 & 91.2 & 86.4 & 82.6 & 95.5 & 70.0 & 93.9 & 84.1 & 55.6 & 75.6 & 82.1\\
3D-GCN \cite{lin2020convolution} & yes & 82.1 & 85.1 & 83.1 & 84.0 & 86.6 & 77.5 & 90.3 & 74.1 & 90.9 & 86.4 & 83.8 & 95.6 & 66.8 & 94.8 & 81.3 & 59.6 & 75.7 & 82.8\\
\ad{HAPGN} \cite{chen2020hapgn} & yes & 87.1 & 89.3 & 87.1 & 85.7 & 90.1 & 86.2 & 91.7 & 78.3 & 94.3 & 85.9 & 82.6 & 95.2 & 77.9 & 94.3 & 90.1 & 73.9 & 90.3 & 90.6\\
\hline
MAP-VAE \cite{han2019multi} & no & 67.95 & - & 62.7 & 67.1 & 73.0 &  58.5 & 77.1 & 67.3 & 84.8 & 77.1 & 60.9 & 90.8 & 35.8 & 87.7 & 64.2 & 45.0 & 60.4 & 74.8\\
Multi-task \cite{hassani2019unsupervised} & no & 72.1 & 77.7 & 78.4 & 67.7 & 78.2 & 66.2 & 85.5 & 52.6 & 87.7 & 81.6 & 76.3 & 93.7 & 56.1 & 80.1 & 70.9 & 44.7 & 60.7 & 73.0\\
\hline
Ours (\tvcjadd{DGCNN}) & no & 75.5 & 79.2 & 76.3 & 76.6 & 82.5 & 65.8 & 85.9 & 67.1 & 86.6 & 81.3 & 79.2 & 93.8 & 55.8 & 92.8 & 73.5 & 53.1 & 61.3 & 76.6\\
    \hline
    \end{tabular}
\end{table*}

\subsection{Scene Segmentation}
\begin{figure}[htb]
\centering
\begin{minipage}[b]{0.48\linewidth}
{\label{}\includegraphics[width=1\linewidth]{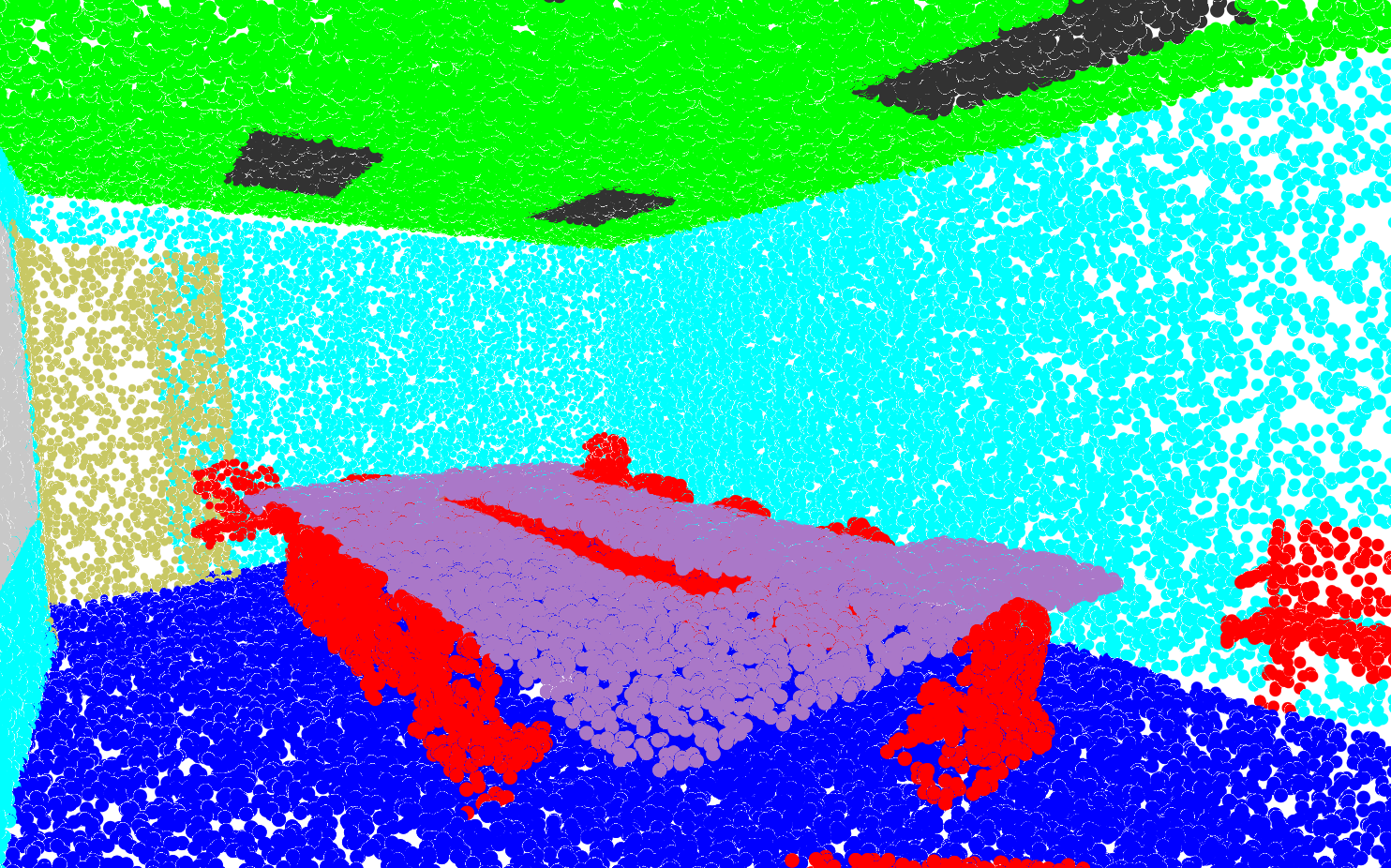}}
\centerline{Ground truth}
\end{minipage}
\begin{minipage}[b]{0.48\linewidth}
{\label{}\includegraphics[width=1\linewidth]{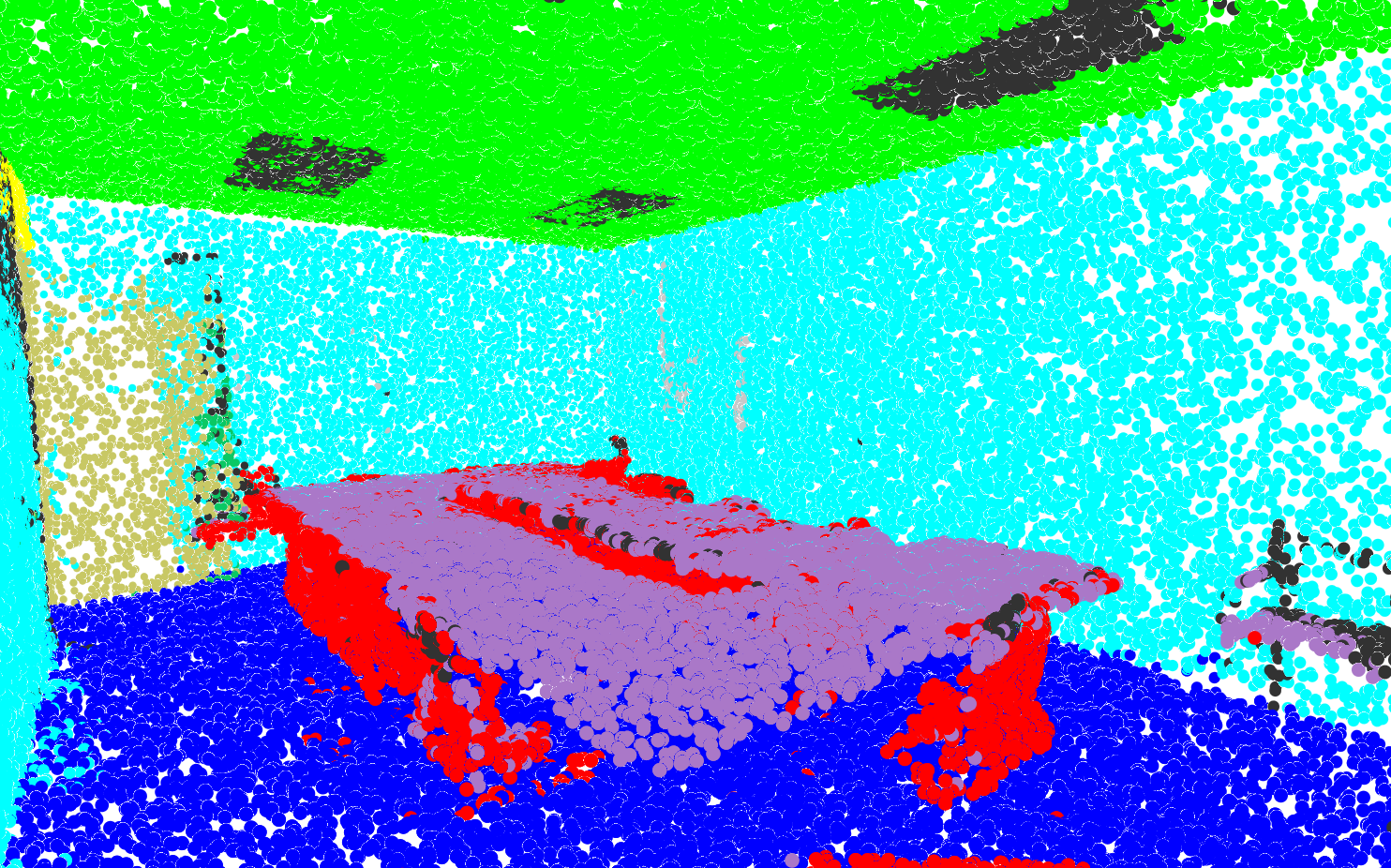}}
\centerline{Our result}
\end{minipage}
\caption{Visual result of scene segmentation.  }
\label{fig:sceneseg}
\end{figure}

We also test our method for the scene segmentation task on the S3DIS  dataset, which typically appears to be more challenging than the shape part segmentation. Similarly, we utilize DGCNN as the backbone and adopt the Linear Classifier evaluation setting. 
We are not able to compare our method with unsupervised methods like MAP-VAE \cite{han2019multi} and Multi-task \cite{hassani2019unsupervised}, since they did not provide scene segmentation results and their source codes are not publicly available. Table \ref{table:scenesegmentationarea5} lists the comparisons of 1 fold testing on  Area 5. It is observed that our method even outperforms the supervised PointNet in terms of mean accuracy. Due to the unsupervised property, our method is inferior to the supervised PointCNN and fine-tuned PointContrast. Our method has relatively smaller mean IOU, which is probably due to the imbalanced categories and the limited minibatch size. The performance could be further improved if more powerful computing resources are allowed.  Figure \ref{fig:sceneseg} shows a visual example for scene segmentation.

\begin{table}[htb]
    \centering
    \caption{Scene segmentation results of our method \tvcjadd{(Linear Classifier)} and some state-of-the-art techniques on testing Area 5 (Fold 1) of the S3DIS dataset. }\label{table:scenesegmentationarea5}
    \begin{tabular}{l c c c}
        \hline
        Methods & Supervised & \tabincell{c}{Mean\\accuracy} & \tabincell{c}{Mean\\IOU}\\
        \hline
        PointNet \cite{qi2017pointnet} & yes & 49.0 & 41.1\\
        \ad{PointCE} \cite{liu2020semantic} & yes & - & 51.7\\
        PointCNN \cite{li2018pointcnn} & yes & 63.9 & 57.3\\
        PointContrast \cite{xie2020pointcontrast} & yes & 76.9 & 70.3\\
        \hline
        Ours (\tvcjadd{DGCNN}) & no & \jc{59.4} & \jc{32.6}\\
    \hline
    \end{tabular}
\end{table}

\subsection{Ablation Studies}
\label{sec:ablation}

\begin{figure}[htb]
\centering
\begin{minipage}[b]{0.24\linewidth}
{\label{}\includegraphics[width=1\linewidth]{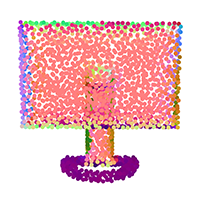}}
\centerline{Original}
\centerline{ }
\end{minipage}
\begin{minipage}[b]{0.24\linewidth}
{\label{}\includegraphics[width=1\linewidth]{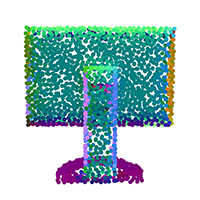}}
\centerline{Rotate $180^\circ$}
\centerline{($Y$ axis)}
\end{minipage}
\begin{minipage}[b]{0.24\linewidth}
{\label{}\includegraphics[width=1\linewidth]{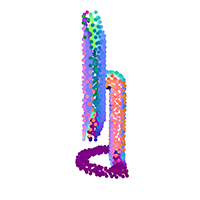}}
\centerline{Rotate $90^\circ$}
\centerline{($Y$ axis)}
\end{minipage} 
\begin{minipage}[b]{0.24\linewidth}
{\label{}\includegraphics[width=1\linewidth]{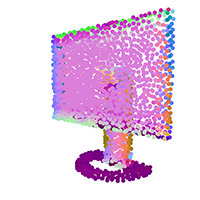}}
\centerline{Rotate $45^\circ$}
\centerline{($Y$ axis)}
\end{minipage}
\\
\begin{minipage}[b]{0.24\linewidth}
{\label{}\includegraphics[width=1\linewidth]{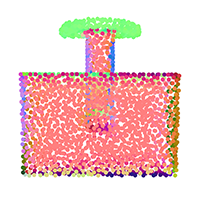}}
\centerline{Rotate $180^\circ$}
\centerline{($X$ axis)}
\end{minipage}
\begin{minipage}[b]{0.24\linewidth}
{\label{}\includegraphics[width=1\linewidth]{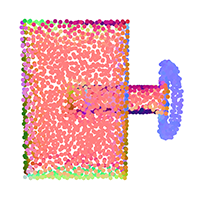}}
\centerline{Rotate $90^\circ$}
\centerline{($X$ axis)}
\end{minipage}
\begin{minipage}[b]{0.24\linewidth}
{\label{}\includegraphics[width=1\linewidth]{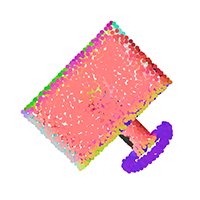}}
\centerline{Rotate $45^\circ$}
\centerline{($X$ axis)}
\end{minipage}
\begin{minipage}[b]{0.24\linewidth}
{\label{}\includegraphics[width=1\linewidth]{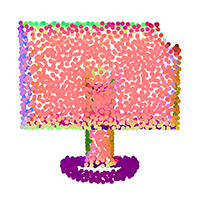}}
\centerline{Cutout}
\centerline{ }
\end{minipage}
\\
\begin{minipage}[b]{0.24\linewidth}
{\label{}\includegraphics[width=1\linewidth]{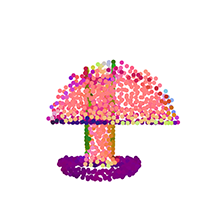}}
\centerline{Crop}
\end{minipage}
\begin{minipage}[b]{0.24\linewidth}
{\label{}\includegraphics[width=1\linewidth]{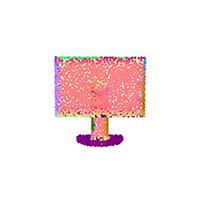}}
\centerline{Scale}
\end{minipage}
\begin{minipage}[b]{0.24\linewidth}
{\label{}\includegraphics[width=1\linewidth]{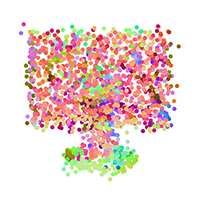}}
\centerline{Jitter}
\end{minipage}
\begin{minipage}[b]{0.24\linewidth}
{\label{}\includegraphics[width=1\linewidth]{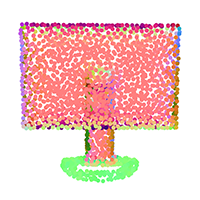}}
\centerline{Smooth}
\end{minipage}
\caption{Illustration for transformations used in Table \ref{table:transformablation}.  }
\label{fig:contrastivetransformation}
\end{figure}

\textbf{Transformation.}
\jc{One of the key elements in our unsupervised representation learning is using $180^\circ$ rotation around the $Y$ axis to get the transformation. To comprehensively study the influence of transformation on representations, we consider many common transformations including rotation, cutout, crop, scale, jittering and smoothing. Figure \ref{fig:contrastivetransformation} visualizes different transformations for a point cloud.}

\jc{We list the comparison results of the above transformations in Table \ref{table:transformablation}. It can be clearly observed that our choice attains the best accuracy, which is unlike SimCLR \cite{chen2020simple} that utilizes two different transformations of an image as the pair. 
We suspect that rotation is a very simple and effective transformation for 3D point cloud data, and a larger valid rotation would generate a greater pose discrepancy (i.e., contrast) in 3D space. As such, our choice using $180^\circ$ rotation around the $Y$ axis is better than others.}

\tvcj{Furthermore, we apply two sequential transformations on one point cloud and make the result as a pair with the original point cloud. We chose the best transformation (i.e. rotate $180^\circ$ around $Y$ axis) as the first transformation, and then apply one of the rest of the transformations as the second. 
For more complex transformations, we group all the transformations into four categories, including rotation, scaling, jittering/smoothing, and cropping/cutout, and apply these four categories on one point cloud sequentially.  
We show the results in Table  \ref{table:transformatwice}. It can be seen that after applying the complex transformation, it is still not as good as the best choice shown above. We suspect that the complex transformation may damage the information on the point cloud, thus leading to inferior results. Again, this verifies that our choice is the best transformation for 3D point cloud data in generating contrastive pairs. 
}

\begin{table}[htb]
    \centering
    \caption{Comparison of different contrastive transformation on ModelNet10. DGCNN \cite{wang2019dynamic} is the backbone. We use linear classification evaluation for comparisons. \% is used for classification accuracy. }\label{table:transformablation}
    \begin{tabular}{l c c}
        \hline
         \tabincell{c}{Transformation} &  \tabincell{c}{Mean class\\accuracy} &  \tabincell{c}{Overall\\accuracy}\\ 
        \hline
             rotate $180^\circ$($Y$ axis) & 94.88 & 95.09\\
             rotate $90^\circ$($Y$ axis) & 94.12 & 94.53\\
             rotate $45^\circ$($Y$ axis) & 94.09 & 94.20\\
             rotate $180^\circ$($X$ axis) & 93.21 & 93.53\\
             rotate $90^\circ$($X$ axis) & 93.30 & 93.42\\
             rotate $45^\circ$($X$ axis) & 93.71 & 93.97\\
             cutout & 94.01 & 93.97\\
             crop & 93.80 & 94.31\\
             scale & 94.10 & 94.20\\
             jitter & 93.95 & 93.97\\
             smooth & 93.93 & 94.08\\
        \hline
    \end{tabular}
\end{table}

\begin{table}[htb]
    \centering
    \caption{\tvcj{Comparison of more complex contrastive transformation on ModelNet10. We distinguish between the results of using only two transformations and those of using four transformations (more complex) by the line. ``Rotate'' means $180^\circ$ rotation around the $Y$ axis. DGCNN \cite{wang2019dynamic} is the backbone. We use linear classification evaluation for comparisons. \% is used for classification accuracy.} }\label{table:transformatwice}
    \begin{tabular}{l c c}
        \hline
         \tabincell{c}{Transformation} &  \tabincell{c}{Mean class\\accuracy} &  \tabincell{c}{Overall\\accuracy}\\ 
        \hline
             rotate + cutout  & 93.43 & 93.64\\
             rotate + crop & 93.90 & 93.97\\
             rotate + scale & 94.11 & 94.08\\
             rotate + jitter & 94.33 & 94.42\\
             rotate + smooth & 93.41 & 93.64\\
        \hline
            rotate + scale + jitter + cutout  & 94.07 & 94.42\\
            rotate + scale + jitter + crop & 93.92 & 94.31\\
            rotate + scale + smooth + cutout  & 93.56 & 93.86\\
            rotate + scale + smooth + crop & 93.78 & 94.20\\
        \hline
             rotate $180^\circ$($Y$ axis) & 94.88 & 95.09\\
        \hline
    \end{tabular}
\end{table}

\textbf{Output of encoder versus output of projection head.}
We also compare the choices of using the output of the base encoder (i.e. global feature) and the output of the projection head for subsequent linear classification. Table \ref{table:encoderhead} shows the comparison results of the two choices on ModelNet40 and ModelNet10. We see that the former choice is better than the latter choice. We think the output of the base encoder involves more discriminative features for the training of the linear classifier.

\begin{table}[htb]
    \centering
    \caption{Comparison of using the output of encoder and projection head for linear classification evaluation. PointNet \cite{qi2017pointnet} is the backbone.
    \% is used for classification accuracy.
    }\label{table:encoderhead}
    \begin{tabular}{l c c c}
        \hline
         \tabincell{c}{Component} &  \tabincell{c}{Dataset} & 
         \tabincell{c}{Mean class\\accuracy} &  \tabincell{c}{Overall\\accuracy}\\ 
        \hline
             encoder & ModelNet40 & 83.81 & 88.65\\
             head & ModelNet40 & 68.55 & 75.81\\
             encoder & ModelNet10 & 90.55 & 90.64\\
             head & ModelNet10 & 81.57 & 82.59\\
        \hline
    \end{tabular}
\end{table}

\textbf{Cross validation.}
In addition to the above evaluations, we further test the abilities of our unsupervised contrastive representation learning in a crossed evaluation setting. To achieve this, we use the learned representations from the unsupervised trained model on ModelNet40 to further train a linear classifier on ModelNet10, and vice versa. Classification outcomes are reported in Table \ref{table:crossvalidation}. It can be observed that our unsupervised representation learning is indeed working in the cross-dataset setting. It also reveals that our unsupervised method trained on a large dataset would probably benefit the testing on another dataset greatly. In here, our method trained on ModelNet40 enables a better cross-test accuracy, compared to unsupervised training on ModelNet10 and testing on ModelNet40. 

\begin{table}[htbp]
    \centering
    \caption{Cross validation for ModelNet40 and ModelNet10. We perform unsupervised learning on one dataset and conduct the classification task on another dataset. PointNet \cite{qi2017pointnet} is the backbone.
    \% is used for classification accuracy.
    }\label{table:crossvalidation}
    \begin{tabular}{c c c c}
        \hline
         \tabincell{c}{Unsupervised\\dataset} &  \tabincell{c}{Classification\\dataset} & 
         \tabincell{c}{Mean class\\accuracy} &  \tabincell{c}{Overall\\accuracy}\\ 
        \hline
             ModelNet40 & ModelNet10 & 90.00 & 90.51\\
             ModelNet10 & ModelNet40 & 77.13 & 82.87\\
        \hline
    \end{tabular}
\end{table}

\textbf{Pretraining evaluation: initializing projection head.}
Projection head is very useful in maximizing the agreement between the contrastive pair. However, it may hinder pretraining evaluation, if the corresponding part is initialized with the projection head of the unsupervised model. Table \ref{table:headablation} shows that initializing encoder only  produces better classification accuracy for PointNet/DGCNN on ModelNet10/ModelNet40, which confirms the judgement that initializing encoder only is a better choice. 

\setlength{\tabcolsep}{3pt}
\begin{table}[htbp]
    \centering
    \caption{Pretraining validation to determine whether using preojection head for initialization. ModelNet40 and ModelNet10 are used for datasets. PointNet \cite{qi2017pointnet} and DGCNN\cite{wang2019dynamic} are the backbone.
    \% is used for classification accuracy.
    }\label{table:headablation}
    \begin{tabular}{c c c c c}
        \hline
         \tabincell{c}{Backbone} &  \tabincell{c}{Dataset} & \tabincell{c}{Head\\initialization} & 
         \tabincell{c}{Mean class\\accuracy} &  \tabincell{c}{Overall\\accuracy}\\ 
        \hline
             PointNet & ModelNet10 & yes & 93.80 & 93.97\\
             PointNet & ModelNet10 & no & 94.23 & 94.38\\
             PointNet & ModelNet40 & yes & 86.64 & 90.22\\
             PointNet & ModelNet40 & no & 86.80 & 90.44\\
             DGCNN & ModelNet10 & yes & 95.05 & 95.09\\
             DGCNN & ModelNet10 & no & 95.78 & 95.93\\
             DGCNN & ModelNet40 & yes & 88.58 & 91.96\\
             DGCNN & ModelNet40 & no & 89.52 & 93.03\\
        \hline
    \end{tabular}
\end{table}

\begin{figure}[htb]
\centering
\begin{minipage}[b]{0.48\linewidth}
{\label{}\includegraphics[width=1\linewidth]{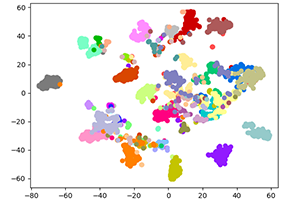}}
\end{minipage}
\begin{minipage}[b]{0.48\linewidth}
{\label{}\includegraphics[width=1\linewidth]{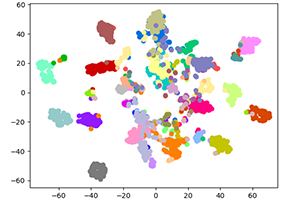}}
\end{minipage}
\caption{\tvcj{T-SNE visualization of features. (a) without contrastive learning, (b) with contrastive learning. } }
\label{fig:tsne}
\end{figure}

\tvcj{
\textbf{T-SNE Visualization.} 
We utilize the t-SNE to visualize the features learned on ModelNet40 in Figure \ref{fig:tsne}. It can be seen that using only a linear classifier can separate different features to some extent. It is worth noting that our model using unsupervised contrastive learning can better separate features after training a linear classifier, which implies that our contrastive learning is useful and  effectively facilitates the classification task.
}

\section{Conclusion}
\label{sec:conclusion}
We have presented an unsupervised representation learning method for 3D point cloud data. We identified that rotation is a very useful transformation for generating a contrastive version of an original point cloud. Unsupervised representations are learned via maximizing the correspondence between paired point clouds (i.e. an original point cloud and its contrastive version). Our method is simple to implement and does not require expensive computing resources like TPU. We evaluate our unsupervised representations for the downstream tasks including 3D object classification, shape part segmentation and scene segmentation. Experimental results demonstrate that our method generates impressive performance. In  the future, We would  like to exploit  semi-supervised techniques like  \cite{feng2020dmt} to improve the performance. We would also like to extend our approach to other interesting applications such as 3D object detection. 

\section*{Conflicts of Interests}
The authors declare that the work is original and has not been submitted elsewhere.

\begin{appendices}

\section{Overview of Segmentation}
\begin{figure*}[htb]\footnotesize
\centering
\begin{minipage}[b]{0.99\linewidth}
{\label{}
\includegraphics[width=1\linewidth]{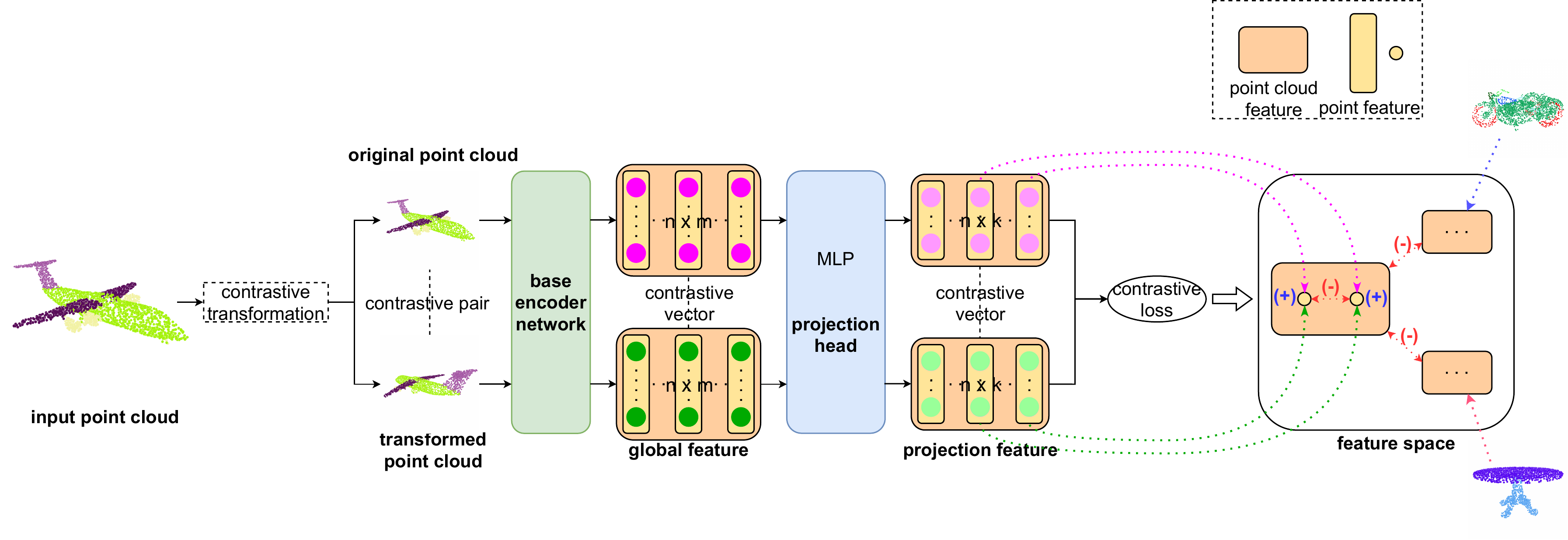}}
\end{minipage}
\caption{ Overview of our unsupervised contrastive learning for the downstream segmentation task. All the point clouds in the minibatch will be mapped into the feature space. The designed contrastive loss (shown in the main paper) encourages a pair of point clouds (original point cloud and its transformed point cloud) to be consistent in  the feature space, and the point-wise features of the same point ID also tend to be consistent. 
}
\label{fig:overview_seg}
\end{figure*}

We also achieve the task of point cloud semantic segmentation, including shape part segmentation and scene segmentation. Different from the 3D object classification task, we need to gain all the point-wise features in the point cloud, which is the key to solve the segmentation task. For our unsupervised contrastive learning, \jc{as shown in Figure \ref{fig:overview_seg}}, we still consider the original point cloud and its transformed point cloud as a contrastive pair.  However, in order to ensure that the feature of each point in the point cloud will be learned, we use the mean of point-wise cross entropy to evaluate the point cloud similarity, and try to maximize the similarity of the positive pair (all other pairs of point clouds in the minibatch are viewed as negative pairs). In this unsupervised manner, our framework can learn the feature of each point in the point cloud.

\section{Additional Visual Results on Scene Segmentation}

\begin{figure}[htb]
\centering
\begin{minipage}[b]{0.45\linewidth}
{\label{}\includegraphics[width=1\linewidth]{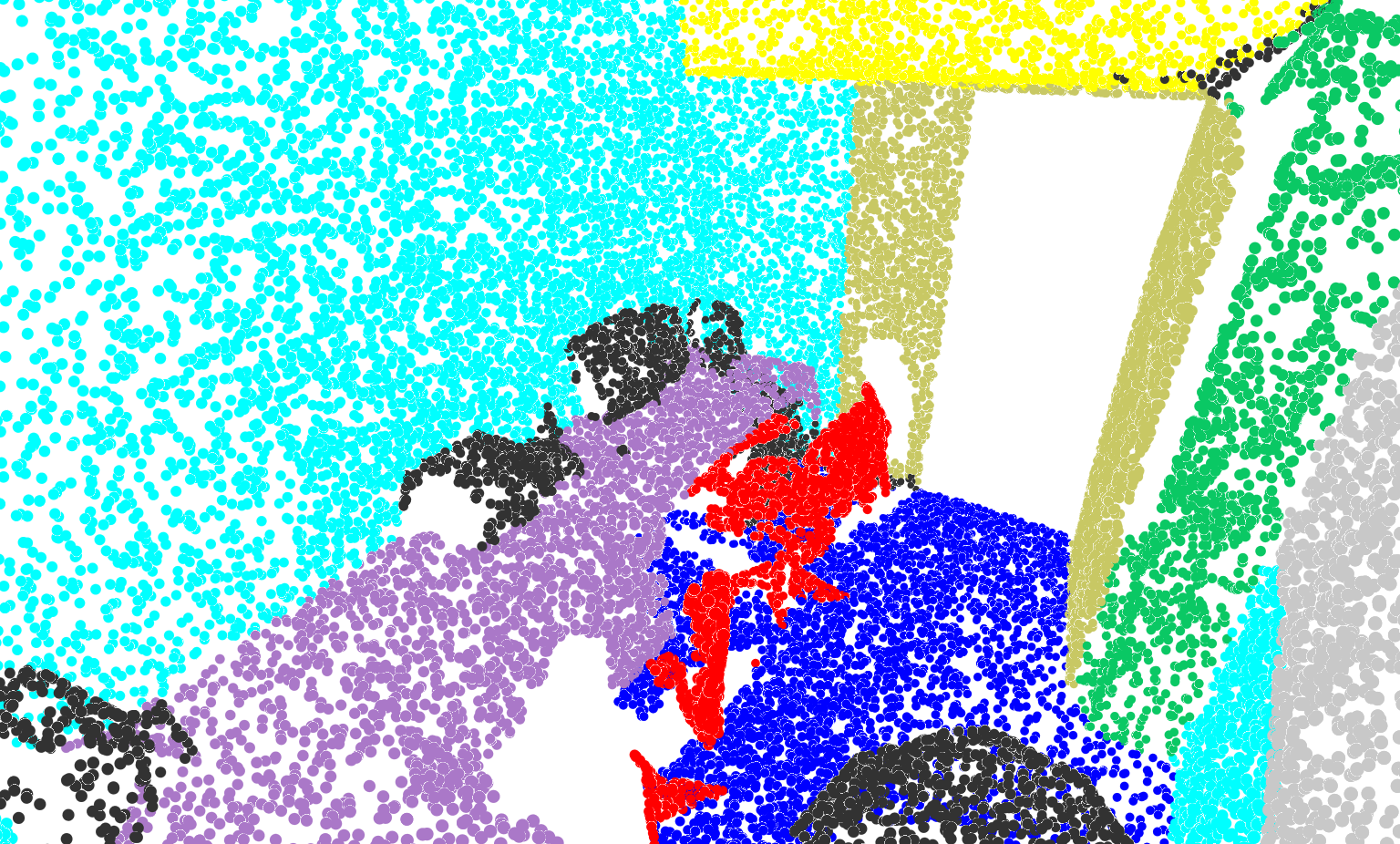}}
\end{minipage}
\hspace{0.3cm}
\begin{minipage}[b]{0.45\linewidth}
{\label{}\includegraphics[width=1\linewidth]{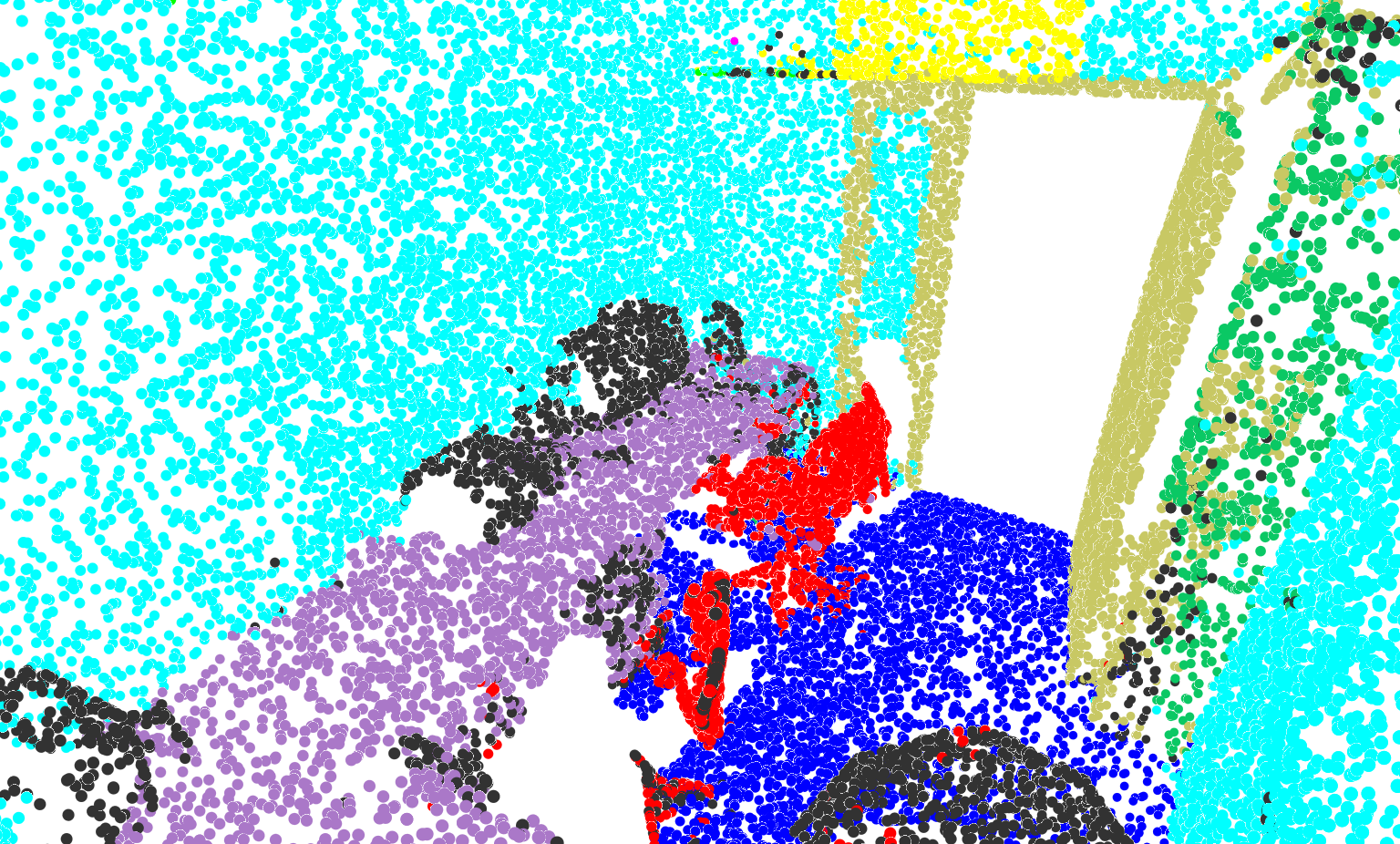}}
\end{minipage}
\\
\vspace{0.5cm}
\centering
\begin{minipage}[b]{0.45\linewidth}
{\label{}\includegraphics[width=1\linewidth]{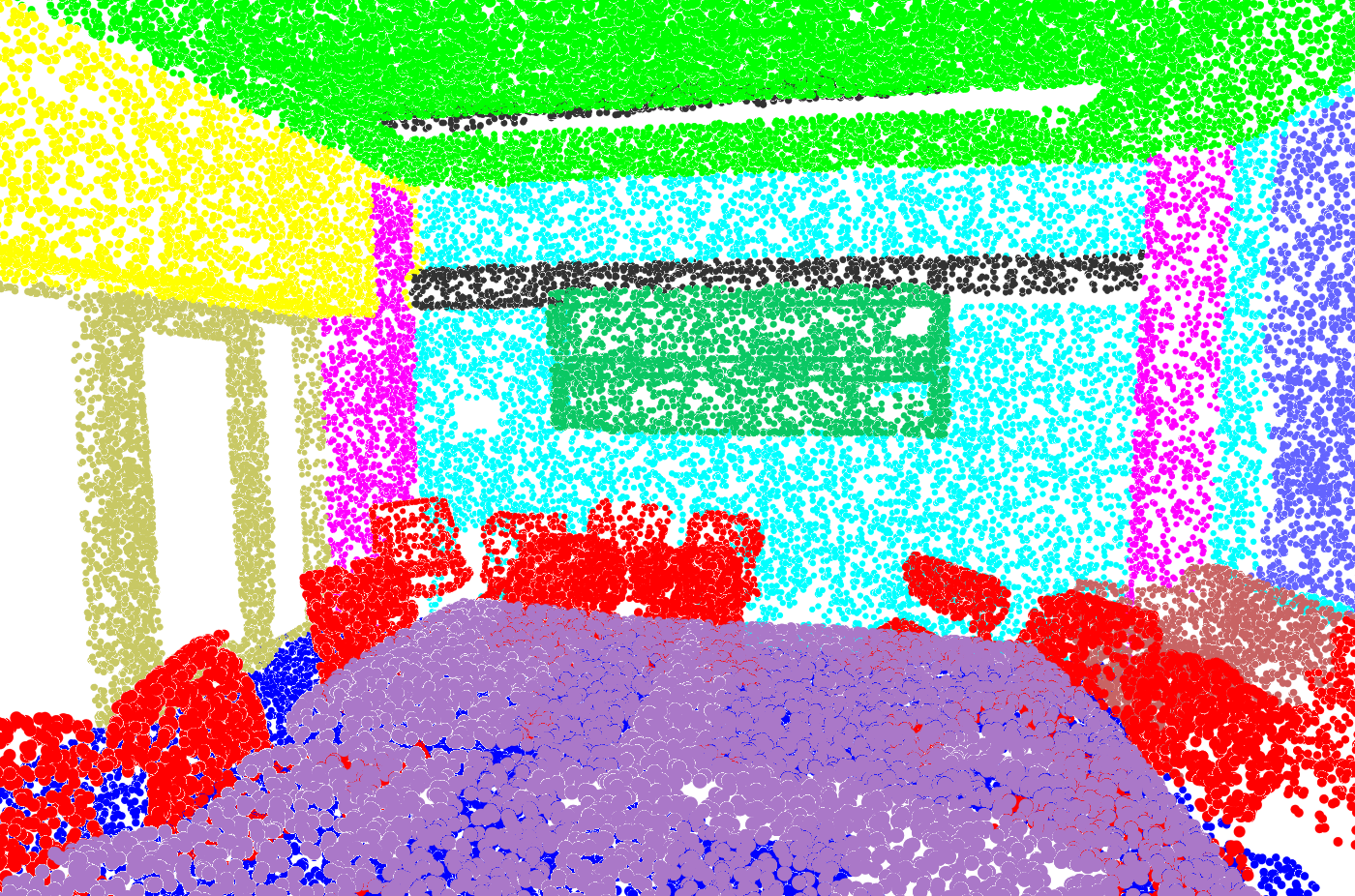}}
\centerline{Ground truth}
\end{minipage}
\hspace{0.3cm}
\begin{minipage}[b]{0.45\linewidth}
{\label{}\includegraphics[width=1\linewidth]{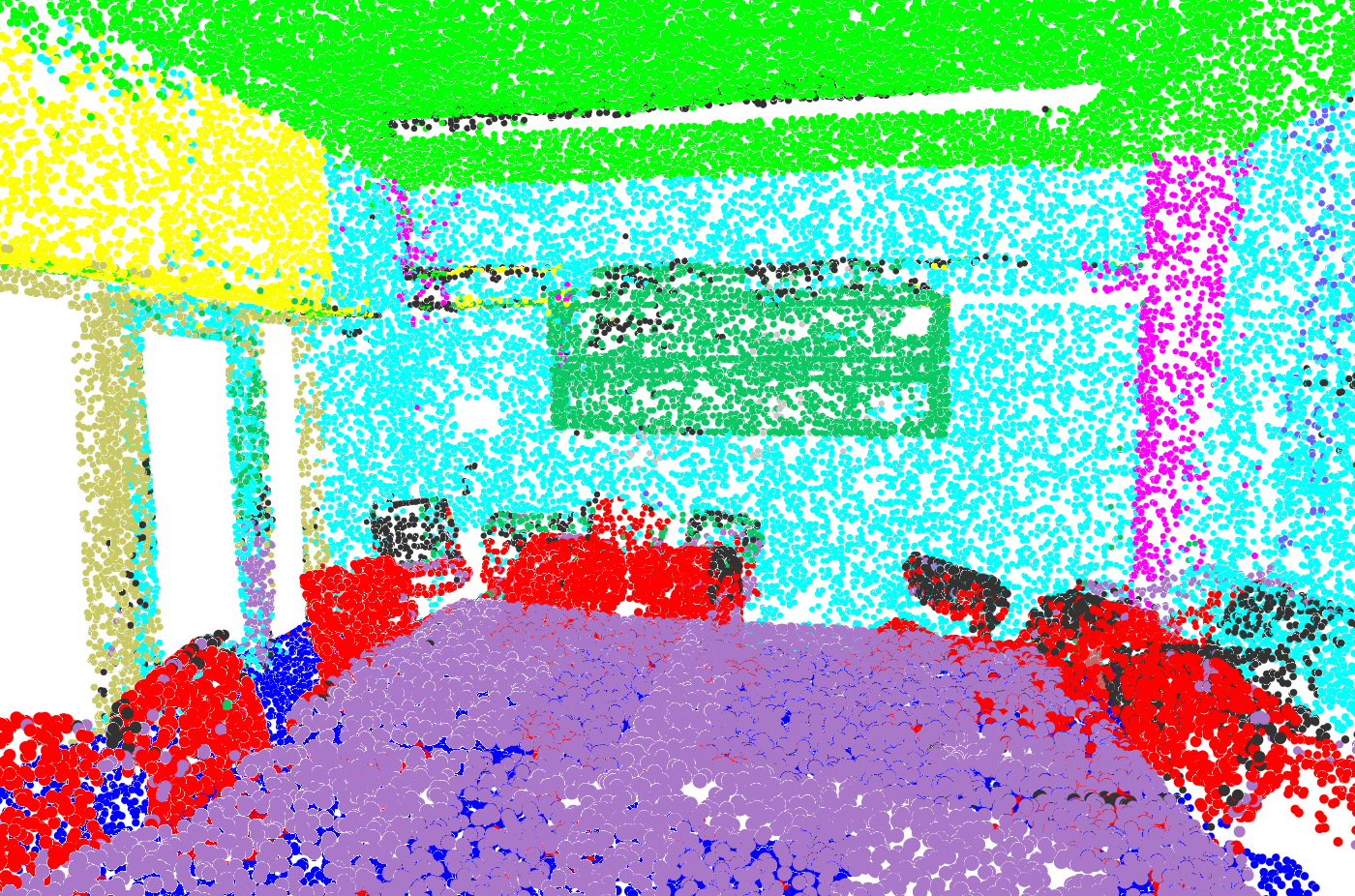}}
\centerline{Our result}
\end{minipage}

\caption{Visual result of scene segmentation.  }
\label{fig:secenesegALL}
\end{figure}

In this section, we show more visual results on scene segmentation. Similarly, we utilize the Linear Classifier setting for this downstream task. Figure \ref{fig:secenesegALL} shows the visual results of several scenes. We can observe from the figure that our method produces close segmentation results to the ground truth. This demonstrates the capability of our unsupervised representation learning method.

\section{Additional Visual Results on Shape Part Segmentation}

\begin{figure*}[htbp]
\centering
\begin{minipage}[b]{0.49\linewidth}
    \begin{minipage}[b]{0.24\linewidth}
    \centerline{Ground truth}
    \end{minipage}
    \begin{minipage}[b]{0.24\linewidth}
    \centerline{Our result}
    \end{minipage}
    \begin{minipage}[b]{0.24\linewidth}
    \centerline{Ground truth}
    \end{minipage}
    \begin{minipage}[b]{0.24\linewidth}
    \centerline{Our result}
    \end{minipage}
\end{minipage}
\begin{minipage}[b]{0.49\linewidth}
    \begin{minipage}[b]{0.24\linewidth}
    \centerline{Ground truth}
    \end{minipage}
    \begin{minipage}[b]{0.24\linewidth}
    \centerline{Our result}
    \end{minipage}
    \begin{minipage}[b]{0.24\linewidth}
    \centerline{Ground truth}
    \end{minipage}
    \begin{minipage}[b]{0.24\linewidth}
    \centerline{Our result}
    \end{minipage}
\end{minipage}
\\
\centering
\begin{minipage}[b]{0.49\linewidth}
    \begin{minipage}[b]{0.24\linewidth}
    {\label{}\includegraphics[width=1\linewidth]{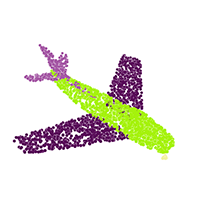}}
    \end{minipage}
    \begin{minipage}[b]{0.24\linewidth}
    {\label{}\includegraphics[width=1\linewidth]{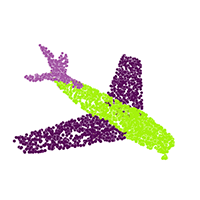}}
    \end{minipage}
    \begin{minipage}[b]{0.24\linewidth}
    {\label{}\includegraphics[width=1\linewidth]{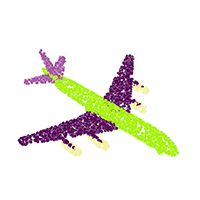}}
    \end{minipage}
    \begin{minipage}[b]{0.24\linewidth}
    {\label{}\includegraphics[width=1\linewidth]{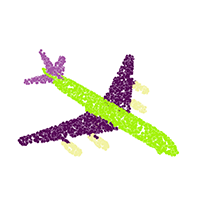}}
    \end{minipage}
\centerline{airplane}
\end{minipage}
\begin{minipage}[b]{0.49\linewidth}
    \begin{minipage}[b]{0.24\linewidth}
    {\label{}\includegraphics[width=1\linewidth]{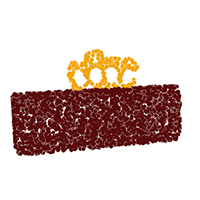}}
    \end{minipage}
    \begin{minipage}[b]{0.24\linewidth}
    {\label{}\includegraphics[width=1\linewidth]{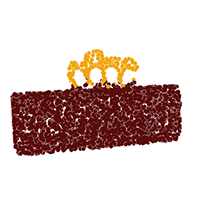}}
    \end{minipage}
    \begin{minipage}[b]{0.24\linewidth}
    {\label{}\includegraphics[width=1\linewidth]{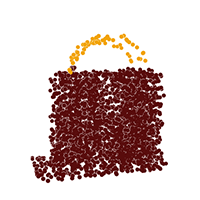}}
    \end{minipage}
    \begin{minipage}[b]{0.24\linewidth}
    {\label{}\includegraphics[width=1\linewidth]{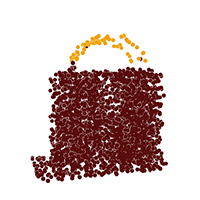}}
    \end{minipage}
\centerline{bag}
\end{minipage}
\\
\centering
\begin{minipage}[b]{0.49\linewidth}
    \begin{minipage}[b]{0.24\linewidth}
    {\label{}\includegraphics[width=1\linewidth]{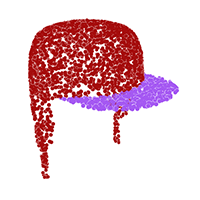}}
    \end{minipage}
    \begin{minipage}[b]{0.24\linewidth}
    {\label{}\includegraphics[width=1\linewidth]{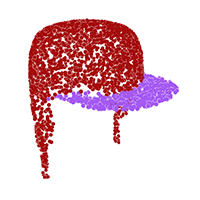}}
    \end{minipage}
    \begin{minipage}[b]{0.24\linewidth}
    {\label{}\includegraphics[width=1\linewidth]{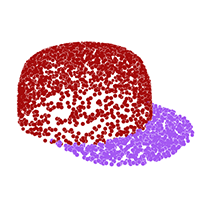}}
    \end{minipage}
    \begin{minipage}[b]{0.24\linewidth}
    {\label{}\includegraphics[width=1\linewidth]{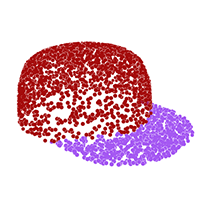}}
    \end{minipage}
\centerline{cap}
\end{minipage}
\begin{minipage}[b]{0.49\linewidth}
    \begin{minipage}[b]{0.24\linewidth}
    {\label{}\includegraphics[width=1\linewidth]{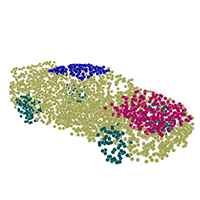}}
    \end{minipage}
    \begin{minipage}[b]{0.24\linewidth}
    {\label{}\includegraphics[width=1\linewidth]{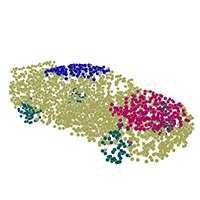}}
    \end{minipage}
    \begin{minipage}[b]{0.24\linewidth}
    {\label{}\includegraphics[width=1\linewidth]{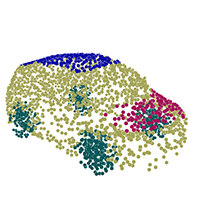}}
    \end{minipage}
    \begin{minipage}[b]{0.24\linewidth}
    {\label{}\includegraphics[width=1\linewidth]{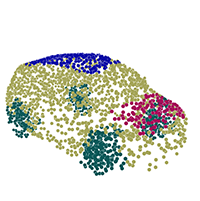}}
    \end{minipage}
\centerline{car}
\end{minipage}
\\
\centering
\begin{minipage}[b]{0.49\linewidth}
    \begin{minipage}[b]{0.24\linewidth}
    {\label{}\includegraphics[width=1\linewidth]{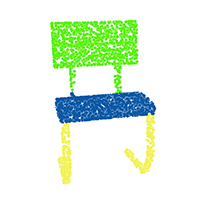}}
    \end{minipage}
    \begin{minipage}[b]{0.24\linewidth}
    {\label{}\includegraphics[width=1\linewidth]{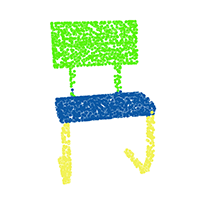}}
    \end{minipage}
    \begin{minipage}[b]{0.24\linewidth}
    {\label{}\includegraphics[width=1\linewidth]{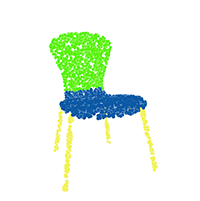}}
    \end{minipage}
    \begin{minipage}[b]{0.24\linewidth}
    {\label{}\includegraphics[width=1\linewidth]{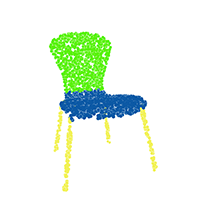}}
    \end{minipage}
\centerline{chair}
\end{minipage}
\begin{minipage}[b]{0.49\linewidth}
    \begin{minipage}[b]{0.24\linewidth}
    {\label{}\includegraphics[width=1\linewidth]{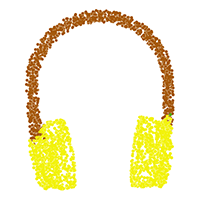}}
    \end{minipage}
    \begin{minipage}[b]{0.24\linewidth}
    {\label{}\includegraphics[width=1\linewidth]{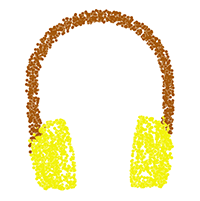}}
    \end{minipage}
    \begin{minipage}[b]{0.24\linewidth}
    {\label{}\includegraphics[width=1\linewidth]{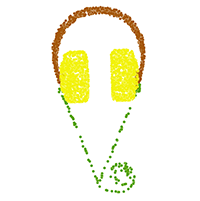}}
    \end{minipage}
    \begin{minipage}[b]{0.24\linewidth}
    {\label{}\includegraphics[width=1\linewidth]{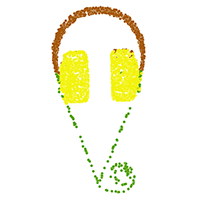}}
    \end{minipage}
\centerline{earphone}
\end{minipage}
\\
\centering
\begin{minipage}[b]{0.49\linewidth}
    \begin{minipage}[b]{0.24\linewidth}
    {\label{}\includegraphics[width=1\linewidth]{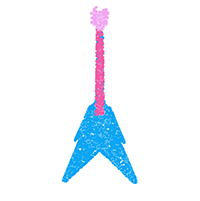}}
    \end{minipage}
    \begin{minipage}[b]{0.24\linewidth}
    {\label{}\includegraphics[width=1\linewidth]{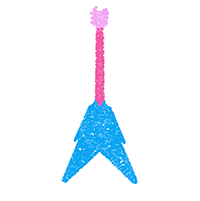}}
    \end{minipage}
    \begin{minipage}[b]{0.24\linewidth}
    {\label{}\includegraphics[width=1\linewidth]{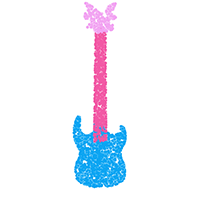}}
    \end{minipage}
    \begin{minipage}[b]{0.24\linewidth}
    {\label{}\includegraphics[width=1\linewidth]{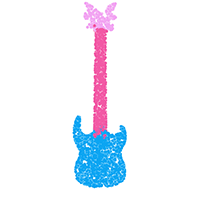}}
    \end{minipage}
\centerline{guitar}
\end{minipage}
\begin{minipage}[b]{0.49\linewidth}
    \begin{minipage}[b]{0.24\linewidth}
    {\label{}\includegraphics[width=1\linewidth]{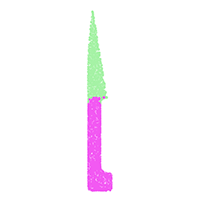}}
    \end{minipage}
    \begin{minipage}[b]{0.24\linewidth}
    {\label{}\includegraphics[width=1\linewidth]{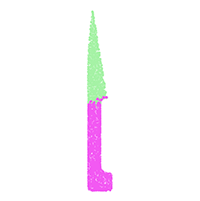}}
    \end{minipage}
    \begin{minipage}[b]{0.24\linewidth}
    {\label{}\includegraphics[width=1\linewidth]{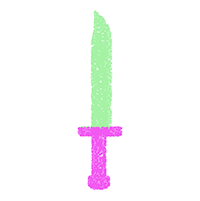}}
    \end{minipage}
    \begin{minipage}[b]{0.24\linewidth}
    {\label{}\includegraphics[width=1\linewidth]{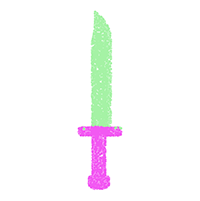}}
    \end{minipage}
\centerline{knife}
\end{minipage}
\\
\centering
\begin{minipage}[b]{0.49\linewidth}
    \begin{minipage}[b]{0.24\linewidth}
    {\label{}\includegraphics[width=1\linewidth]{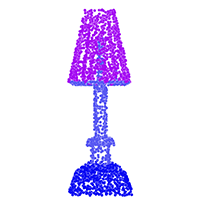}}
    \end{minipage}
    \begin{minipage}[b]{0.24\linewidth}
    {\label{}\includegraphics[width=1\linewidth]{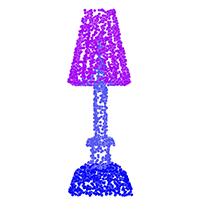}}
    \end{minipage}
    \begin{minipage}[b]{0.24\linewidth}
    {\label{}\includegraphics[width=1\linewidth]{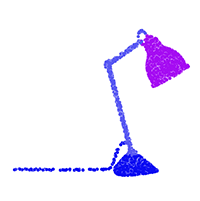}}
    \end{minipage}
    \begin{minipage}[b]{0.24\linewidth}
    {\label{}\includegraphics[width=1\linewidth]{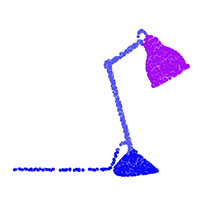}}
    \end{minipage}
\centerline{lamp}
\end{minipage}
\begin{minipage}[b]{0.49\linewidth}
    \begin{minipage}[b]{0.24\linewidth}
    {\label{}\includegraphics[width=1\linewidth]{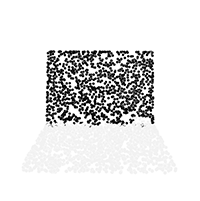}}
    \end{minipage}
    \begin{minipage}[b]{0.24\linewidth}
    {\label{}\includegraphics[width=1\linewidth]{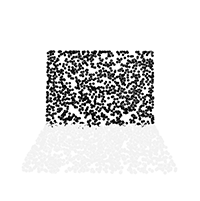}}
    \end{minipage}
    \begin{minipage}[b]{0.24\linewidth}
    {\label{}\includegraphics[width=1\linewidth]{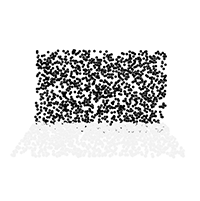}}
    \end{minipage}
    \begin{minipage}[b]{0.24\linewidth}
    {\label{}\includegraphics[width=1\linewidth]{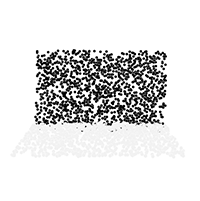}}
    \end{minipage}
\centerline{laptop}
\end{minipage}
\\
\centering
\begin{minipage}[b]{0.49\linewidth}
    \begin{minipage}[b]{0.24\linewidth}
    {\label{}\includegraphics[width=1\linewidth]{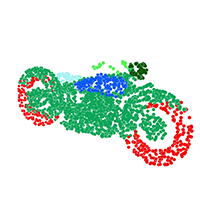}}
    \end{minipage}
    \begin{minipage}[b]{0.24\linewidth}
    {\label{}\includegraphics[width=1\linewidth]{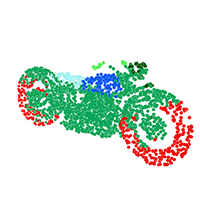}}
    \end{minipage}
    \begin{minipage}[b]{0.24\linewidth}
    {\label{}\includegraphics[width=1\linewidth]{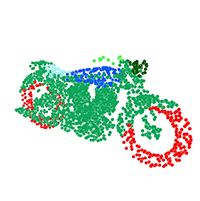}}
    \end{minipage}
    \begin{minipage}[b]{0.24\linewidth}
    {\label{}\includegraphics[width=1\linewidth]{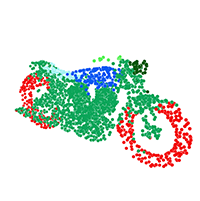}}
    \end{minipage}
\centerline{motorbike}
\end{minipage}
\begin{minipage}[b]{0.49\linewidth}
    \begin{minipage}[b]{0.24\linewidth}
    {\label{}\includegraphics[width=1\linewidth]{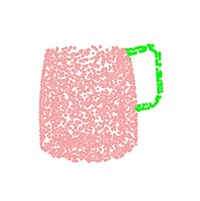}}
    \end{minipage}
    \begin{minipage}[b]{0.24\linewidth}
    {\label{}\includegraphics[width=1\linewidth]{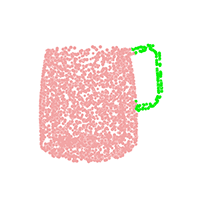}}
    \end{minipage}
    \begin{minipage}[b]{0.24\linewidth}
    {\label{}\includegraphics[width=1\linewidth]{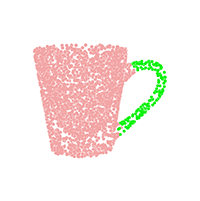}}
    \end{minipage}
    \begin{minipage}[b]{0.24\linewidth}
    {\label{}\includegraphics[width=1\linewidth]{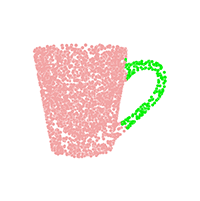}}
    \end{minipage}
\centerline{mug}
\end{minipage}
\\
\centering
\begin{minipage}[b]{0.49\linewidth}
    \begin{minipage}[b]{0.24\linewidth}
    {\label{}\includegraphics[width=1\linewidth]{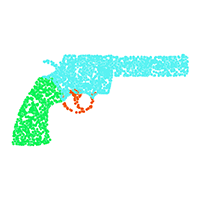}}
    \end{minipage}
    \begin{minipage}[b]{0.24\linewidth}
    {\label{}\includegraphics[width=1\linewidth]{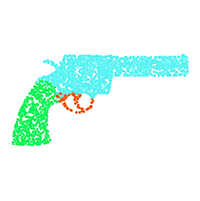}}
    \end{minipage}
    \begin{minipage}[b]{0.24\linewidth}
    {\label{}\includegraphics[width=1\linewidth]{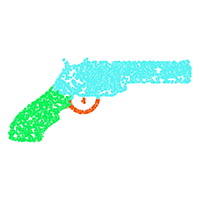}}
    \end{minipage}
    \begin{minipage}[b]{0.24\linewidth}
    {\label{}\includegraphics[width=1\linewidth]{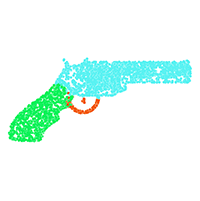}}
    \end{minipage}
\centerline{pistol}
\end{minipage}
\begin{minipage}[b]{0.49\linewidth}
    \begin{minipage}[b]{0.24\linewidth}
    {\label{}\includegraphics[width=1\linewidth]{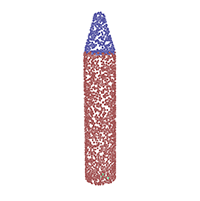}}
    \end{minipage}
    \begin{minipage}[b]{0.24\linewidth}
    {\label{}\includegraphics[width=1\linewidth]{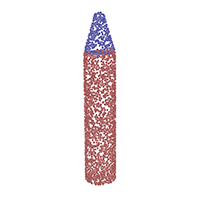}}
    \end{minipage}
    \begin{minipage}[b]{0.24\linewidth}
    {\label{}\includegraphics[width=1\linewidth]{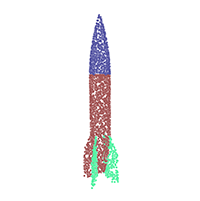}}
    \end{minipage}
    \begin{minipage}[b]{0.24\linewidth}
    {\label{}\includegraphics[width=1\linewidth]{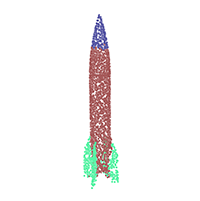}}
    \end{minipage}
\centerline{rocket}
\end{minipage}
\\
\centering
\begin{minipage}[b]{0.49\linewidth}
    \begin{minipage}[b]{0.24\linewidth}
    {\label{}\includegraphics[width=1\linewidth]{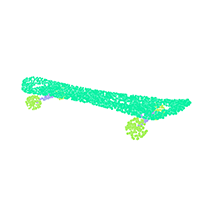}}
    \end{minipage}
    \begin{minipage}[b]{0.24\linewidth}
    {\label{}\includegraphics[width=1\linewidth]{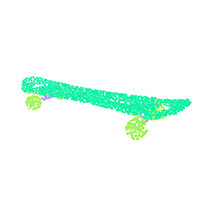}}
    \end{minipage}
    \begin{minipage}[b]{0.24\linewidth}
    {\label{}\includegraphics[width=1\linewidth]{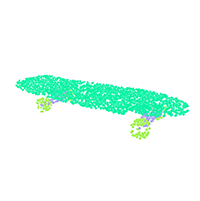}}
    \end{minipage}
    \begin{minipage}[b]{0.24\linewidth}
    {\label{}\includegraphics[width=1\linewidth]{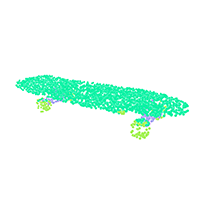}}
    \end{minipage}
\centerline{skateboard}
\end{minipage}
\begin{minipage}[b]{0.49\linewidth}
    \begin{minipage}[b]{0.24\linewidth}
    {\label{}\includegraphics[width=1\linewidth]{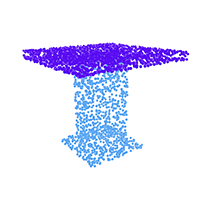}}
    \end{minipage}
    \begin{minipage}[b]{0.24\linewidth}
    {\label{}\includegraphics[width=1\linewidth]{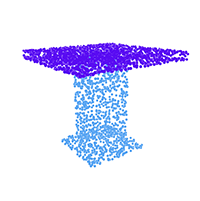}}
    \end{minipage}
    \begin{minipage}[b]{0.24\linewidth}
    {\label{}\includegraphics[width=1\linewidth]{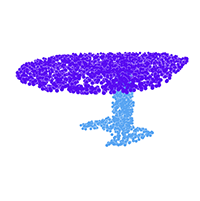}}
    \end{minipage}
    \begin{minipage}[b]{0.24\linewidth}
    {\label{}\includegraphics[width=1\linewidth]{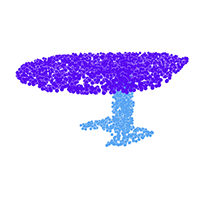}}
    \end{minipage}
\centerline{table}
\end{minipage}

\caption{Some examples of all 16 categories in ShapeNet Part dataset. }
\label{fig:partsegall}
\end{figure*}

In this section, we put more visual results of our method on the downstream shape part segmentation. We simply employ the Linear Classifier setting for this downstream task. Figure \ref{fig:partsegall} shows the visual results of 32 models of 16 categories, involving 2 models per category. As we can see from the figure, with our unsupervised learned representations, a simple linear classifier for the downstream task can generate very similar visual results to the ground truth segmentation. It further confirms the effectiveness of our unsupervised method in learning distinguishable representations.

\end{appendices}


\newpage

\bibliography{sn-bibliography}


\end{document}